\documentclass[10pt,journal,compsoc]{IEEEtran}

\ifCLASSOPTIONcompsoc
  \usepackage[nocompress]{cite}
\else
  \usepackage{cite}
\fi

\ifCLASSINFOpdf
\else
\fi

\usepackage{graphicx}
\usepackage[pdftex, colorlinks,linkcolor={red},anchorcolor={blue},citecolor={green},urlcolor={magenta}, backref=page]{hyperref}
\usepackage{hyperref}

\usepackage{booktabs}
\usepackage{multirow}
\usepackage{pifont} 

\usepackage{graphicx}
\usepackage{subcaption}
\usepackage{amsmath}

\usepackage{amsfonts}

\usepackage{color}
\usepackage{xcolor}
\definecolor{lightblue}{rgb}{0.12, 0.60, 0.92} 
\definecolor{lightpurple}{rgb}{0.66, 0.29, 0.86} 

\begin{document}

\title{Towards Ultrafast Depth Sensing Via Active Event-based Stereo Vision}

\author{Jianing~Li,~\IEEEmembership{Member,~IEEE},
        Yunjian~Zhang,
        Haiqian~Han,
        Kangyao~Huang,
        and~Xiangyang~Ji,~\IEEEmembership{Member,~IEEE}
\IEEEcompsocitemizethanks{
\IEEEcompsocthanksitem Jianing Li is with the School of Computer Science, Peking University, Beijing 100871, China, and also with Peng Cheng Laboratory, Shenzhen 518000, China (e-mail: lijianing@pku.edu.cn).
\IEEEcompsocthanksitem Yunjian Zhang, Haiqian Han, and Xiangyang Ji are with the Department of Automation, Tsinghua University, Beijing 100084,  Beijing, P.R. China. E-mail: lijianing@pku.edu.cn, sdtczyj@gmail.com, hanhq23@mails.tsinghua.edu.cn, xyji@tsinghua.edu.cn.
\IEEEcompsocthanksitem Kangyao Huang is with the Department of Computer Science and Technology, Tsinghua University, Beijing 100084, P.R. China. E-mail: kangyao.huang@outlook.com.
\IEEEcompsocthanksitem Manuscript received September 7, 2025; revised January 15, 2026; accepted March 8, 2026.
\IEEEcompsocthanksitem (Corresponding author: Xiangyang Ji.)}}

\markboth{IEEE Transactions on Pattern Analysis and Machine Intelligence}%
{Li \MakeLowercase{\textit{et al.}}: Towards Ultrafast Depth Sensing Via Active Event-based Stereo Vision}

\IEEEtitleabstractindextext{%
\begin{abstract}
Conventional frame-based imaging for active stereo systems has encountered major challenges in fast-motion scenarios. However, how to design a novel paradigm for ultrafast depth sensing remains an open issue. In this paper, we propose a novel problem setting, namely active event-based stereo vision, which attempts to integrate binocular event cameras and an infrared 2D pattern projector for high-speed dense depth sensing. Technically, we first build a stereo camera prototype system and present a real-world dataset with over 21.5k spatiotemporal synchronized labels at 15 Hz, while also establishing a realistic synthetic dataset with stereo event streams and 23.8k synchronized labels at 20 Hz. Then, we propose ActiveEventNet+, a lightweight yet effective event-based stereo matching neural network that learns to generate high-quality dense disparity maps from stereo event streams with low latency. Our ActiveEventNet+ mainly involves three innovations: incorporating lightweight blocks into event-based stereo matching frameworks, designing a novel cost volume with dynamic interactions between stereo pairs, and presenting an effective temporal consistency architecture to fully use rich temporal cues in event streams. The results show that our ActiveEventNet+ outperforms state-of-the-art methods while significantly reducing computational complexity. Our solution offers superior depth sensing performance compared to conventional frame-based stereo cameras in high-speed scenes. In particular, the lightweight ActiveEventNet enables the prototype system to achieve real-time processing at speeds up to 150 FPS. We believe that this novel active event-based stereo vision paradigm can provide new insights into the design of future high-speed depth sensing camera systems. Our dataset and code can be available at \url{https://github.com/jianing-li/active\_event\_based\_stereo}.
\end{abstract}

}

\maketitle

\IEEEdisplaynontitleabstractindextext

\IEEEpeerreviewmaketitle

\IEEEraisesectionheading{\section{Introduction} \label{sec:introduction}}

\IEEEPARstart{S}{tereo} depth perception~\cite{laga2020survey, poggi2021synergies}, one of the longstanding and fundamental topics, supports a wide range of computer vision and robotics tasks. Passive stereo vision typically struggles in texture-less regions and low-light environments~\cite{jeon2016stereo}. In contrast, active stereo vision~\cite{fanello2017ultrastereo} addresses these challenges by projecting an infrared pattern, enabling more accurate depth maps compared to passive stereo systems. Nevertheless, conventional active stereo cameras are generally constrained by their depth frame rates (e.g., 30 FPS for Kinect V1 and 90 FPS for RealSense D435), limiting their effectiveness in high-speed scenarios~\cite{loquercio2021learning, falanga2020dynamic}. For instance, a racing drone could suffer a severe collision in a short period between two adjacent depth frames. This may raise a key question: \emph{How can we develop a novel active stereo sensing paradigm for high-speed depth perception to overcome the limitations of conventional stereo cameras?}

Event cameras~\cite{lichtsteiner2008128, gallego2020event}, also known as silicon retinas, operate differently from conventional frame-based cameras. Instead of capturing frames at a fixed rate, they detect changes in intensity at each pixel, generating asynchronous events with microsecond-level temporal resolution~\cite{li2021asynchronous}. This capability makes them ideal for various high-speed vision tasks~\cite{rebecq2019high, wan2022learning, liu2025high, muglikar2023event, bao2024temporal, he2024microsaccade, sun2024unified, yin2024exploring, sun2022event, liu2025eventbench, yu2025evagaussians, liu2025eventgpt} that require low-latency processing. Consequently, there is growing research interest in leveraging event cameras for high-speed depth sensing (i.e., monocular and binocular) in agile robots~\cite{falanga2019fast}. In particular, event-based stereo vision~\cite{zhu2018realtime, nam2022stereo, zhou2018semi, tulyakov2019learning} offers the advantage of providing more accurate and reliable depth information compared to monocular depth estimation~\cite{liu2024event, zhu2019unsupervised, liu2022event, chiavazza2023low, li2025asynchronous}.

One problem is that current event-based stereo depth systems~\cite{zhang2022discrete, uddin2022unsupervised, cho2023learning, andreopoulos2018low, ahmed2021deep, liu2022learning, jiang2025event, ghosh2025depth} are passive stereo, resulting in \emph{inaccurate depth maps in texture-less regions and dark scenes}. In other words, passive event-based stereo methods rely on feature matching, which may be challenging or even impossible in areas with little texture or low contrast. While some structured light systems with a single event camera~\cite{muglikar2021esl, bajestani2023event, muglikar2021event, huang2021high, wang2022enhancing, wang2020temporal, li2024event, fu2023fast, yu2025active} have attempted to achieve high-speed depth sensing, these active monocular depth methods may not always match the accuracy of stereo depth systems in challenging scenes. This limitation exists because monocular systems depend entirely on the precise detection of projected patterns, which can be easily compromised by highly absorptive or texture-less surfaces. Conversely, a binocular system benefits from the fusion of active and passive perception. By integrating active infrared patterns with passive environmental textures from two viewpoints, the system leverages strong geometric constraints through stereo triangulation to ensure robust depth estimation across diverse real-world settings. Yet, the integration of binocular event cameras with structured light in active stereo systems has received relatively limited exploration. Meanwhile, existing passive event-based stereo vision datasets~\cite{zhu2018multivehicle, chaney2023m3ed, gehrig2021dsec} typically provide only sparse LiDAR depth maps, and there is a lack of event streams with structured light and dense depth labels.

Another problem is that most event-based stereo matching methods~\cite{tulyakov2019learning,  mostafavi2021event, uddin2022unsupervised, liu2022learning, nam2022stereo, cho2022selection, ranccon2022stereospike, cho2023learning, chen2024event} prioritize maximizing accuracy through more complex feed-forward neural networks, leading to \emph{high computational complexity and wasted rich temporal cues}. This may be contrary to the advantages of event cameras, which offer low latency and rich temporal data. For example, some stereo matching algorithms~\cite{zhu2018realtime, uddin2022unsupervised, nam2022stereo} aim to build higher-dimensional architectures (e.g., 3D or 4D cost volumes) to further enhance accuracy, yet few efforts have been made to develop lightweight models that explicitly optimize computational efficiency and inference speed. Moreover, these methods typically process event bins independently using feed-forward models, thereby discarding valuable temporal cues from continuous event streams. While leveraging temporal information can improve accuracy, it often comes at the cost of increased computational overhead. Thus, a key challenge lies in designing an efficient event-based stereo matching model that achieves a balance between accuracy and speed while fully utilizing rich temporal cues.

\begin{figure}[t]
\centering
\includegraphics[width=\linewidth]{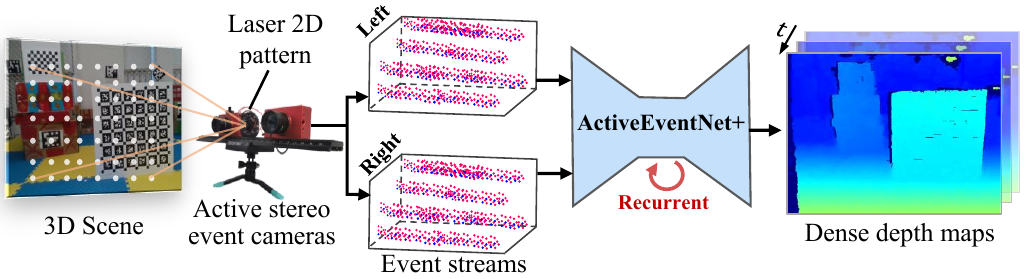}
\caption{Our active event-based stereo camera system integrates binocular event cameras and an infrared 2D pattern laser for high-speed depth sensing. Our ActiveEventNet+ efficiently generates dense depth maps with low latency, achieving an optimal accuracy-speed trade-off while fully exploiting rich temporal cues from stereo event streams.}
\label{fig:motivation}
\vspace{-0.20cm}
\end{figure}

To address the aforementioned problems, we propose a novel paradigm for ultrafast dense depth sensing, namely active event-based stereo vision, which attempts to integrate binocular event cameras and an infrared 2D pattern projector to enable low-latency dense depth estimation (see Fig.~\ref{fig:motivation}). The goal of this work is to develop and validate a novel active event-based stereo vision system that addresses the following challenges: (i) \emph{Lack of prototype system and dataset} – How do we establish an active stereo camera setup and build high-quality event-based datasets, including both simulated and real-world scenarios? (ii) \emph{Lightweight yet effective model} – How can we design a lightweight event-based stereo matching model that balances accuracy and speed while effectively leveraging rich temporal cues?

To this end, we first build an active event-based stereo camera prototype and present a real-world dataset with over 21.5k spatiotemporal synchronized true depth labels at 15 Hz, while also creating a highly realistic synthetic dataset with 23.8k synchronized labels at 20 Hz. Then, we design a lightweight yet effective event-based stereo matching neural network (i.e., \textbf{ActiveEventNet+}), which generates high-quality dense disparity maps from stereo event streams with low latency. The proposed ActiveEventNet+ incorporates three key innovations: (i) Pioneering the integration of lightweight MobileNet blocks into event-based stereo matching modeling; (ii) Designing a novel cost volume with dynamic interactions between stereo event stream pairs; and (iii) Devising an efficient temporal consistency architecture to fully utilize rich temporal cues in stereo event streams. Experimental results demonstrate that our ActiveEventNet+ outperforms state-of-the-art stereo matching methods while significantly reducing computational complexity. Our prototype system, integrating binocular event cameras and an infrared projector, delivers superior depth sensing performance compared to conventional frame-based stereo cameras in high-speed motion scenarios. In particular, our lightweight ActiveEventNet enables the prototype system to achieve real-time processing at speeds up to 150 FPS. Furthermore, our solution also highlights that active event-based stereo vision surpasses passive stereo in low-texture regions and dark environments. We believe that this novel active event-based stereo vision paradigm can provide new insights into the development of next-generation neuromorphic stereo camera systems.

In summary, the main contributions of this work are:
\begin{itemize}
\item We present a novel problem setting, termed \emph{active event-based stereo vision}, which attempts to integrate binocular event cameras and an infrared 2D pattern projector for high-speed dense depth sensing.
\item We propose ActiveEventNet+, a lightweight yet effective event-based stereo matching model, which achieves an optimal speed-accuracy trade-off by efficiently utilizing rich temporal cues in event streams.
\item We establish a \emph{real-world} dataset using our active event-based stereo camera prototype, along with a highly \emph{realistic synthetic} dataset containing temporally continuous labels. We believe these open-source standardized datasets will open up new opportunities for research in this novel problem setting.
\end{itemize} 

A preliminary conference version of ActiveEventNet~\cite{li2025active} has appeared in CVPR 2025, which has demonstrated the effectiveness of active event-based stereo vision for high-speed depth sensing. Compared to the prior conference version, this long article is improved in the following aspects: (i) Our ActiveEventNet+ enhances the preliminary model (i.e., ActiveEventNet) by incorporating an efficient temporal consistency architecture, which leverages rich temporal cues in event streams to optimally balance accuracy and inference speed. (ii) \emph{We present more statistical details on two standardized datasets and conduct extensive experiments to verify the effectiveness of our ActiveEventNet+.} (iii) \emph{We propose an extension of a joint framework that effectively leverages the complementary strengths of events and frames for stereo matching, achieving superior performance over unimodal frame-based methods in low-light, low-texture, and high-speed motion scenarios.}

The rest of the paper is organized as follows. Section~\ref{sec:related_work} reviews related work. In Section~\ref{sec:active_stereo_dataset}, we describe our stereo camera prototype and two newly constructed datasets. Section~\ref{sec:preliminary_definition} presents the event camera's working principle and formulates the problem setting. Section~\ref{sec:method} details our lightweight yet effective event-based stereo matching model. Section~\ref{sec:experiment} analyzes the proposed method's performance, while also presenting an extension of the joint framework that integrates stereo events and frames. Finally, Section~\ref{sec:discussion} and Section~\ref{sec:conclusion} provide discussion and conclusions.

\section{Related Work} \label{sec:related_work}
This section first reviews conventional stereo cameras, then surveys neuromorphic stereo datasets and methods, and finally covers neuromorphic vision with structured light.

\subsection{Conventional Stereo Camera Systems} \label{subsec:conventional_stereo}
Conventional stereo cameras~\cite{laga2020survey, poggi2021synergies} can be broadly categorized into passive and active stereo systems. Passive stereo cameras (e.g., ZED~\cite{abdelsalam2024depth}) typically rely on hand-crafted matching algorithms to establish correspondences between two RGB images. However, passive stereo systems face significant challenges in texture-less regions and low-light environments, where the lack of sufficient texture information hinders effective feature matching. Active stereo camera systems~\cite{grosso1995active, fanello2017ultrastereo, zhang2018activestereonet} extend the capabilities of passive stereo by projecting a pseudorandom pattern into the scene using an infrared light source. By selecting the appropriate sensing wavelength, active stereo cameras capture a combination of structured active illumination and ambient passive light, enabling robust depth sensing in both indoor and outdoor environments. For example, Intel R200~\cite{kuan2019comparative} is one of the commercially available active stereo sensors, featuring binocular cameras with a resolution of 640$\times$480 and capable of producing depth maps at 60 FPS. This is followed by Microsoft Kinect V1~\cite{yang2015evaluating}, which captures depth images at 30 FPS. Note that, the Intel D400 series~\cite{keselman2017intel} further enhances these capabilities, providing high-resolution depth maps with a maximum frame rate of 90 FPS. Despite the widespread commercialization of conventional frame-based stereo cameras for various vision tasks, they remain limited by their depth frame rates. This constraint reduces their effectiveness in real-world high-speed applications, particularly those requiring low-latency depth sensing~\cite{falanga2019fast}, such as high-speed drone obstacle avoidance. Thus, this work aims to design a novel event-based stereo vision solution for high-speed depth sensing.

\subsection{Neuromorphic Stereo Datasets} \label{subsec:stereo_dataset}
Several efforts~\cite{ghosh2024event, steffen2019neuromorphic} have been made to develop multi-camera systems with stereo event cameras and LiDAR for unmanned systems, creating multi-modal datasets (e.g., MVSEC~\cite{zhu2018multivehicle}, DSEC~\cite{gehrig2021dsec}, VECtor~\cite{gao2022vector}, M3ED~\cite{chaney2023m3ed}, FusionPortable~\cite{jiao2022fusionportable, wei2024fusionportablev2}, CoSEC~\cite{peng2024cosec}, and PKU-Spike-Stereo~\cite{wang2022learning}) for various tasks. Notably, MVSEC~\cite{zhu2018multivehicle} is a pioneering event-based stereo dataset featuring two binocular DAVIS cameras and a LiDAR sensor for ground truth depth. PKU-Spike-Stereo~\cite{wang2022learning} is the first high-speed depth estimation dataset utilizing two spike cameras~\cite{huang20231000, zhu2022ultra} alongside a ZED camera. DSEC~\cite{gehrig2021dsec} is a widely used multi-modal dataset supporting various vision tasks in autonomous driving scenarios. While publicly available passive stereo datasets using neuromorphic cameras are gradually increasing, active event-based stereo datasets remain largely unexplored. Besides, existing passive event-based stereo vision datasets typically provide only sparse LiDAR depth maps. Therefore, this work aims to establish an active event-based stereo matching dataset with dense depth labels.

\subsection{Neuromorphic Stereo Matching Approaches} \label{subsec:event_stereo}
Since Misha Mahowald's groundbreaking work~\cite{mahowald1992vlsi} in 1992 on the first neuromorphic stereo vision system, the field~\cite{ghosh2024event} has rapidly advanced with the commercialization of neuromorphic cameras. Existing neuromorphic stereo matching methods can broadly be categorized into two categories. The first category refers to traditional model-based methods~\cite{schraml2015event, xie2017event, zhou2018semi, zhu2018realtime, andreopoulos2018low, risi2021instantaneous, shiba2024secrets}, which estimate per-pixel disparity from stereo event streams using reliable local correspondences and global optimization algorithms. While these model-based methods can generate sparse or semi-dense depth maps with low latency, they may be hard to obtain globally dense depth maps. The second category is end-to-end event-based stereo matching networks~\cite{nam2022stereo, tulyakov2019learning,  mostafavi2021event, ahmed2021deep, ranccon2022stereospike, liu2022learning, zhang2022discrete, uddin2022unsupervised, cho2022selection, cho2023learning, chen2024event, cho2025temporal, bartolomei2025lidar, jiang2024ev, jiang2025event, zhang2025enhanced, zhang2025ematch, jiang2025event}, which have become the mainstream approach and have achieved state-of-the-art performance. For example, early attempts (e.g., DDES~\cite{tulyakov2019learning}) typically convert stereo events into image-like representation pairs, enabling compatibility with frame-based neural networks. Subsequently, joint frameworks~\cite{mostafavi2021event, nam2022stereo, uddin2022unsupervised, cho2022selection} are developed to integrate frames and events, aiming to obtain high-quality dense maps. There have also been explorations of hybrid stereo matching strategies~\cite{zuo2021accurate, wang2021stereo} that use a frame-based camera and an event camera to form stereo pairs for evaluating disparity and depth. While existing methods achieve reasonable depth estimation from stereo event streams, they have two key drawbacks. On one hand, their improved accuracy comes at a high computational cost, requiring substantial GPU memory that limits deployment on resource-constrained systems. On the other hand, they usually process each event bin independently using simple feed-forward models, wasting rich temporal cues. This leads to poor depth estimation performance in low-texture or out-of-focus scenarios. Such heavy stereo matching architectures contradict event cameras' advantages of low latency and rich temporal cues. Thus, we propose an efficient event-based stereo matching framework that preserves low latency while fully exploiting temporal cues.


\begin{table*}
\caption{Comparison with existing neuromorphic vision systems using various structured light sources. Note that, our work is the first to explore an active event-based stereo vision system utilizing end-to-end learning-based methods for dense depth sensing in high-speed scenarios.}
\label{tab:related_work_survey}
\vspace{-0.35cm}
\begin{center}
    \renewcommand{\arraystretch}{1.05}
    \setlength{\tabcolsep}{0.90mm}{
        \begin{tabular}{l cccc ccccc}
            \toprule
            Camera system & Venue & Year & Type & Camera & Resolution & Projector & Dataset label & Method  & Density   \\
            \hline
            Brandli \emph{et al.}~\cite{brandli2014adaptive} & FNS & 2014 & Monocular & DVS128 & 128$\times$128 & Laser line, 500 Hz & No & Model-based & Sparse \\
            Leroux \emph{et al.}~\cite{leroux2018event} &  arXiv & 2018 & Monocular & ATIS & 304$\times$260 & Laser 2D pattern & No & Model-based & Sparse \\
            Martel \emph{et al.}~\cite{martel2018active} & ISCAS & 2018 & Stereo & DAVIS240 & 240$\times$180 & Laser point, laser line & No & Model-based & Sparse \\
            FFP~\cite{mangalore2020neuromorphic} & SPL & 2020 & Monocular &  DAVIS346 & 346$\times$260 & Laser 2D pattern & No & Model-based & Sparse \\
            Muglikar \emph{et al.}~\cite{muglikar2021esl}  & 3DV & 2021 & Monocular & Gen3 & 640$\times$480 & Laser point, 60 Hz & No & Model-based & Sparse \\
            Muglikar \emph{et al.}~\cite{muglikar2021event} & 3DV & 2021 & Monocular & Gen3 & 640$\times$480 & Laser point, 60 Hz & Dense & Learning-based & Dense \\
            Huang \emph{et al.}~\cite{huang2021high} & OE & 2021 & Monocular & CeleX-V & 1280$\times$800 & Laser 2D pattern, 9,500 Hz & No & Model-based & Sparse \\
            Takatani \emph{et al.}~\cite{takatani2021event} & CVPR & 2021 & Monocular & DAVIS346 & 346$\times$260 & Laser beam & No & Model-based & Sparse \\
            Wang \emph{et al.}~\cite{wang2022enhancing} & MTF & 2022 & Monocular & DVXplorer & 640$\times$480 & Laser 2D pattern & Dense & Learning-based & Dense \\
            Morgenstern \emph{et al.}~\cite{morgenstern2023x} & CVPRW & 2023 & Monocular & Gen3 & 640$\times$480 & Laser line, 60 Hz & No & Model-based & Sparse \\
            Bajestani \emph{et al.}~\cite{bajestani2023event} & WACV & 2023 & Monocular & Gen3 & 640$\times$480 & Laser 2D pattern, 4,225 Hz & No & Model-based & Sparse \\
            SEpi-3D~\cite{yang2023sepi} & OE & 2023 & Monocular & Gen3 & 640$\times$480 & Laser 2D patterns, 60 Hz & Semi-dense & Model-based & Sparse \\
            Fu \emph{et al.}~\cite{fu2023fast} & OE & 2023 & Monocular & EKV4 & 1280$\times$720 & Laser 2D patterns, 60 Hz & No & Model-based & Sparse \\
            Sirikonda \emph{et al.}~\cite{sirikonda2024structured} & arXiv & 2024 & Monocular & EKV4 & 1280$\times$720 & Laser line, 2M Hz & No & Model-based & Sparse \\
            eFPSL~\cite{li2024event} & SJ & 2024 & Monocular &  DAVIS346 & 346$\times$260 & Laser 2D pattern, 60 Hz & Dense & Model-based & Sparse \\
            SGE~\cite{lu2024sge} & OE & 2024 & Monocular & EKV4 & 1280$\times$720 & Laser line, 60 Hz & No & Model-based & Sparse \\ \hline
            \textbf{Our AESVS} & - & 2024 & Stereo & DAVIS346 & 346$\times$260 & Laser 2D pattern & Dense & Learning-based & Dense \\
            \bottomrule
    \end{tabular}}
\end{center}
\vspace{-0.30cm}
\end{table*}

\subsection{Neuromorphic Vision with Structured Light} \label{subsec:event_structured_light}
Neuromorphic cameras are increasingly being utilized in combination with infrared structured light for high-speed 3D sensing~\cite{guo2024eventlfm} (see Table~\ref{tab:related_work_survey}). In general, structured light sources are commonly categorized into three types (i.e., point, line, and 2D pattern). For instance, Brandli \emph{et al.}~\cite{brandli2014adaptive} first integrates a laser line-projector and an event camera for 3D reconstruction in high-speed scenarios. Martel \emph{et al.}~\cite{martel2018active} combine a laser light source with an event-based stereo setup to quickly scan a scene. Muglikar \emph{et al.}~\cite{muglikar2021esl} design a structured light system with a laser point-projector and an event camera for depth estimation. Huang \emph{et al.}~\cite{huang2021high} present a structured light system using an event camera and a laser pattern-projector for high-speed 3D scanning. Lu \emph{et al.}~\cite{lu2024sge} design a structured light system using gray codes and an event camera, achieving a depth acquisition speed of over 1,000 Hz. Indeed, most active event-based vision systems~\cite{brandli2014adaptive, leroux2018event, martel2018active, muglikar2021esl, huang2021high, wang2020temporal, takatani2021event, mangalore2020neuromorphic, bajestani2023event, morgenstern2023x, yang2023sepi, dashpute2023event, fu2023fast, li2024event, fujimoto2022structured, sirikonda2024structured, lu2024sge, fu2025event} using traditional model-based methods achieve sparse or semi-dense with low latency but lack density. To obtain dense depth maps, several end-to-end learning-based methods~\cite{muglikar2021event, wang2022enhancing} have been developed specifically for event-based depth estimation with structured light. For example, Muglikar \emph{et al.}~\cite{muglikar2021event} propose a novel bio-inspired event-guided depth estimation method using deep neural networks for an event-based structured light system. Wang \emph{et al.}~\cite{wang2022enhancing} present a multi-modal feature fusion network to obtain high-quality depth maps for a hybrid structured light imaging system. While monocular active event-based vision systems are simpler in design, stereo vision systems achieve higher depth estimation accuracy by leveraging triangulation to capture spatial information more effectively. In this work, we design an active event-based stereo vision system, which uses an end-to-end learning-based model for dense depth sensing.

\section{Active Event-based Stereo Dataset} \label{sec:active_stereo_dataset}
This section first describes the construction of our Active-Event-Stereo dataset using our camera prototype and provides statistics for better understanding.

\subsection{Stereo Camera Prototype and Calibration} \label{subsec:camera_prototype}
\begin{figure}[t]
	\begin{subfigure}[b]{0.49\linewidth}
		\centering
		\centerline{\includegraphics[width=4.20cm]{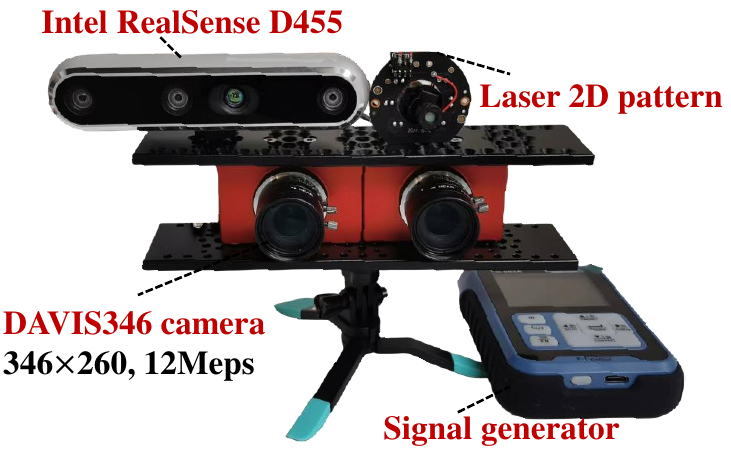}}
		\subcaption{Stereo camera prototype}
		\label{fig:system_dataset(a)}
	\end{subfigure}
	\begin{subfigure}[b]{0.49\linewidth}
		\centering
		\centerline{\includegraphics[width=4.20cm]{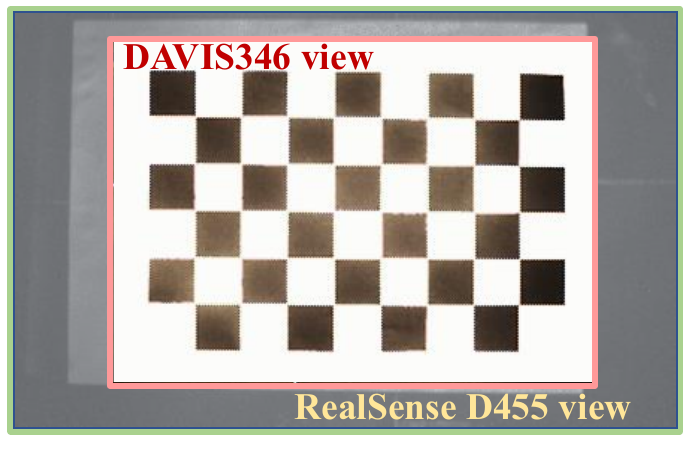}}
		\subcaption{Calibrated views}
		\label{fig:system_dataset(b)}
	\end{subfigure}
	\caption{An active event-based stereo camera prototype. (a) The experimental setup combines binocular event cameras and an infrared pattern projector. (b) Spatiotemporal calibration between two cameras using a standard checkerboard.}
	\label{fig:camera_prototype}
    \vspace{-0.30cm}
\end{figure}

To verify the effectiveness of our solution, we develop an active event-based stereo vision system (AESVS) by integrating binocular DAVIS346 cameras (resolution 346×260), an infrared 2D pattern projector, and an Intel RealSense D455 camera (resolution 640×480). Unlike conventional passive vision, our stereo camera prototype can detect dynamic events even in static scenarios or dark environments by adjusting the laser frequency or intensity of the signal generator. As shown in Fig.~\ref{fig:camera_prototype}(\subref{fig:system_dataset(a)}), our system utilizes an infrared 2D pattern projector with a wavelength of 850 nm, a power of 360 mW, and a field of view of 91° × 65°. While the projector's modulation frequency is adjustable up to 20,000 Hz, it is fixed at 60 Hz during dataset acquisition. Besides, we employ the RealSense D455 stereo camera at 15 FPS to capture depth ground truth in normal motion scenes and provide a fair comparison in high-speed motion scenarios. 

In general, spatiotemporal calibration is a critical step for hybrid multi-camera systems~\cite{li2022retinomorphic}. For \emph{temporal calibration}, we synchronize the two stereo event cameras and the RealSense D455 camera by publishing each topic's timestamp in the robot operating system (ROS). For \emph{spatial calibration}, one objective is to achieve horizontal baseline correction for accurate stereo matching between the two event cameras. Another goal is to align the RealSense D455 camera’s view with that of the left DAVIS346 camera. As illustrated in Fig.~\ref{fig:camera_prototype}(\subref{fig:system_dataset(b)}), a standard checkerboard is positioned in front of our active event-based stereo camera system to provide a complete view. We first use a professional binocular stereo matching correction toolbox to perform baseline correction on the RGB images captured by the two DAVIS346 cameras. Meanwhile, checkerboard keypoints are extracted from the RGB images of both the left DAVIS346 camera and the RealSense D455 camera, and an affine transformation toolbox is applied to align the two coordinate sets. After spatiotemporal synchronization, all sequences will be manually reviewed by a team composed of multiple members.

\begin{figure}[t]
	\centerline{\includegraphics[width=\linewidth]{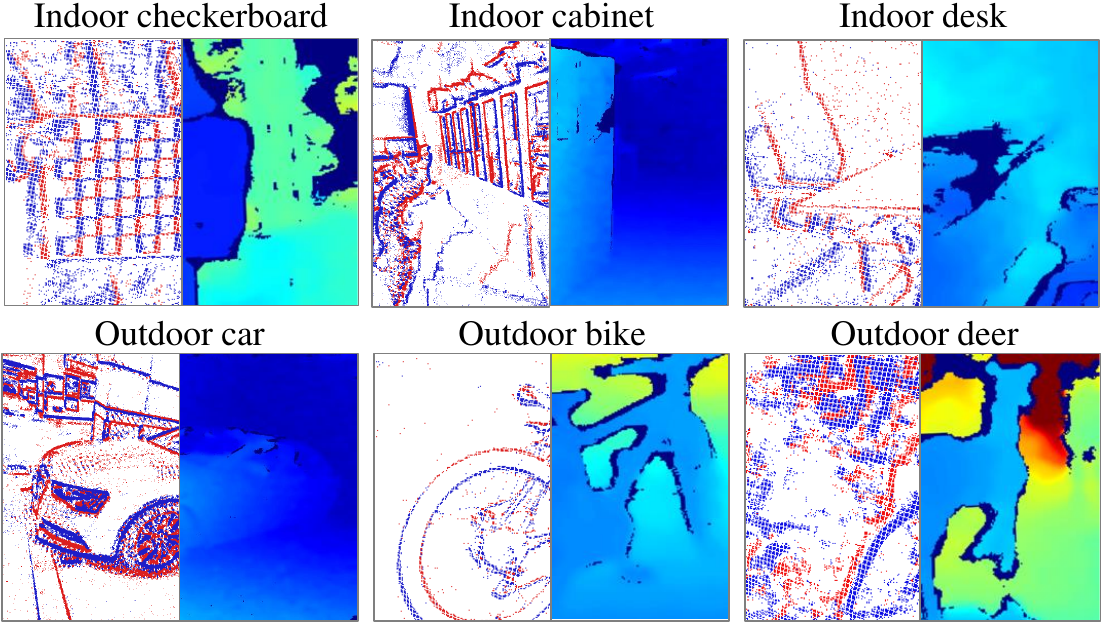}}
	\caption{Representative spatiotemporal synchronized examples. Indoor scenarios are showcased in the top row, and outdoor scenarios in the bottom row. Each pair shows the left DAVIS346 view and the RealSense D455 depth map.} 
	\label{fig:dataset_examples}
\end{figure}

\subsection{Dataset Recording and Statistics} \label{subsec:real_world_dataset}
To provide a high-quality and challenging benchmark, our Active-Event-Stereo dataset considers factors such as velocity distribution, illumination changes, scene diversity, and varying distances during data collection in indoor and outdoor scenarios. Utilizing our stereo camera prototype, we capture 85 sequences that include synchronized event stream pairs, RGB frames, infrared frames, and depth values. After spatiotemporal calibration, the dataset includes ground truth depth labels generated at a frequency of 15 Hz from the RealSense D455. As illustrated in Fig.~\ref{fig:dataset_examples}, we showcase some representative spatiotemporal synchronized examples from our Active-Event-Stereo dataset, showcasing both indoor and outdoor scenarios. Furthermore, we summarize the number of depth labels in each set in Table~\ref{tab:stereo_dataset}. The newly built dataset includes stereo-paired event streams and 21.5k synchronized ground truth labels. These depth labels are subsequently divided into 14.6k for training, 3.6k for validation, and 3.3k for testing. In addition, we further analyze the attributes of our dataset from four perspectives.

\begin{table}
    \begin{center}
    \caption{The number of ground truth depth labels in each set of our newly built Active-Event-Stereo dataset.}
    \renewcommand{\arraystretch}{1.05}
    \label{tab:stereo_dataset}
    \setlength{\tabcolsep}{1.80mm}{
        \begin{tabular}{l cc cc cc}
            \toprule
            \multirow{2}*{Name} & \multicolumn{2}{c}{Indoor scenes} & \multicolumn{2}{c}{Outdoor scenes} \\ 
            \cline{2-3} \cline{4-5} & Normal light & Low-light  & Normal light & Low-light  \\ 
            \hline
            Train & 6,736 & 2,944 & 2,848 & 2,115 \\
            Validation & 1,953 & 829 & 620 & 199 \\
            Test & 983 & 1651 & 419 & 247 \\
            \hline
            Total & 9,654 & 5,424 & 3,887 & 2,561 \\
            \bottomrule
    \end{tabular}}
    \end{center}
    \vspace{-0.30cm}
\end{table}

\begin{figure}[t]
    \begin{subfigure}[b]{0.49\linewidth}
        \centering
        \centerline{\includegraphics[height=2.52cm]{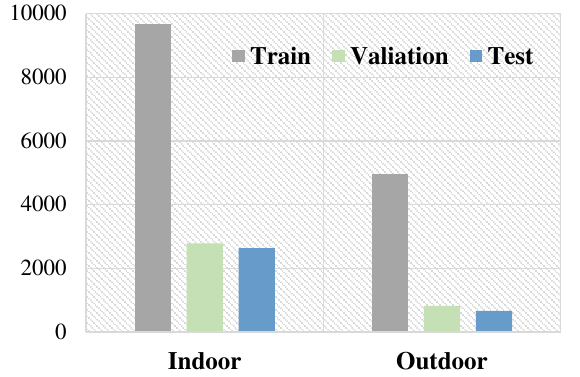}}
        \caption{Scene distribution}
        \label{fig:data_statistics(a)}
    \end{subfigure}
    \begin{subfigure}[b]{0.49\linewidth}
        \centering
        \centerline{\includegraphics[height=2.48cm]{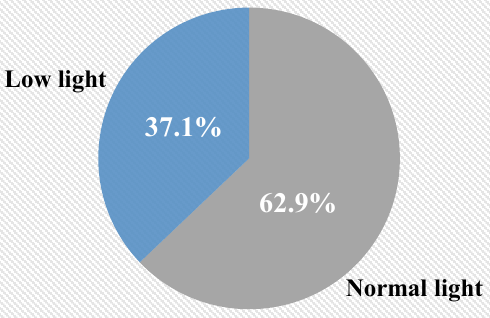}}
        \caption{Light condition}
        \label{fig:data_statistics(b)}
    \end{subfigure}
    \vspace{0.20cm}
    
    \begin{subfigure}[b]{0.49\linewidth}
        \centering
        \centerline{\includegraphics[height=2.48cm]{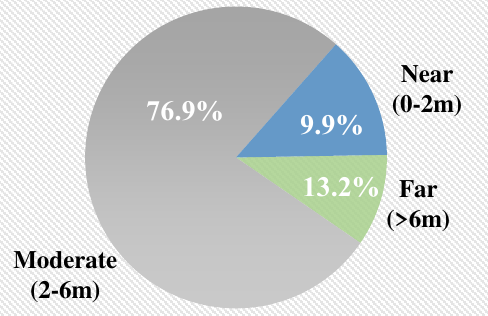}}
        \caption{Distance distribution}
        \label{fig:data_statistics(c)}
    \end{subfigure}
    \begin{subfigure}[b]{0.49\linewidth}
        \centering
        \centerline{\includegraphics[height=2.48cm]{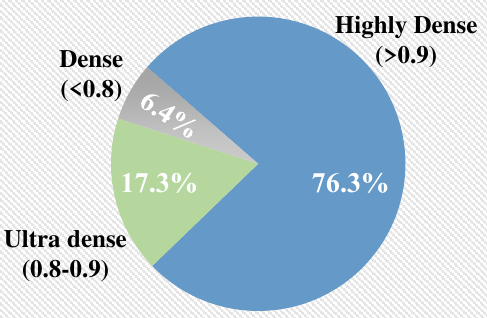}}
        \caption{Depth Density}
        \label{fig:data_statistics(d)}
    \end{subfigure}
    \caption{Dataset statistics in our Active-Event-Stereo dataset. (a) The distribution of indoor and outdoor scenes shows the balance between the two categories. (b-d) The proportions of light conditions, working distances, and depth densities.}
    \label{fig:data_statistics}
    \vspace{-0.30cm}
\end{figure}

\emph{Scene Diversity}. To ensure comprehensive coverage of diverse scenarios, we record sequences from various indoor scenes (e.g., checkerboards, cabinets, and desks) and outdoor scenarios (e.g., cars, bikes, deer, and plants). Fig.~\ref{fig:dataset_examples} shows representative visualization samples from indoor and outdoor scenes. Fig.~\ref{fig:data_statistics}(\subref{fig:data_statistics(a)}) illustrates the distribution of depth labels for indoor and outdoor scenes. More precisely, the number of indoor samples in the training, validation, and testing sets is (9,680, 2,782, 2,634), while the corresponding numbers for outdoor samples are (4,963, 819, 666).

\emph{Light Condition}. To validate the effectiveness of the event-based structured light system under low-light conditions, we ensure that our dataset covers a wide range of illumination conditions from daytime to nighttime. The low-light scenarios primarily include indoor low-light environments and outdoor nighttime scenes. As the DAVIS346 camera provides both RGB and event streams, we can easily identify low-light conditions by visually inspecting the RGB frames. As shown in Fig.~\ref{fig:data_statistics}(\subref{fig:data_statistics(b)}), the proportions of normal light and low-light samples are 62.9\% and 37.1\%, respectively.

\emph{Distance Variation}. To comprehensively cover the working distance range of our stereo camera prototype, we collect indoor and outdoor scenes at various distances. The average depth value of a scene typically reflects the working distance of the camera system. We broadly classify all scenes into three scales (i.e., near, medium, and far). The medium scale is preliminarily defined as ranging from 2m to 6m, with near distances referring to average depth values less than 2m, and far distances corresponding to average depth values greater than 6m. As shown in Fig.~\ref{fig:data_statistics}(\subref{fig:data_statistics(c)}), the proportions of scene average distances in the near, medium, and far scales are 9.9\%, 76.9\%, and 13.3\%, respectively.

\emph{Depth Density}. To quantitatively assess the density of depth labels, we categorize them into three levels (i.e., highly dense, ultra dense, and dense) by the proportion of valid depth pixels relative to the total pixels. Specifically, depth labels are considered highly dense if the valid depth pixel ratio exceeds 0.9, ultra dense if the ratio falls between 0.8 and 0.9, and dense if the ratio is below 0.8. As illustrated in Fig.~\ref{fig:data_statistics}(\subref{fig:data_statistics(d)}), the proportions of depth labels in the highly dense, ultra dense, and dense levels are 76.3\%, 17.3\%, and 6.4\%, respectively. This result shows that the depth labels in our dataset are predominantly dense, providing high-quality ground truth for dense depth estimation.

All in all, such a novel active event-based stereo camera system with structured light and professional design enables our Active-Event-Stereo to be a competitive dataset with \emph{multiple unique characteristics}: (i) \emph{High temporal resolution from event streams with millisecond-level latency}; (ii) \emph{Dynamic event generation with infrared structured light even in static scenes or dark environments}; (iii) \emph{Temporally long-term stereo event stream pairs with dense depth labels at 15 Hz}; (iv) \emph{Real-world recordings with abundant diversities in scene categories, moving speed, light changes, and distance variations}.

\section{Preliminary and Problem Definition} \label{sec:preliminary_definition}
This section first explains how DAVIS cameras work under structured light conditions, then defines the problem of active event-based stereo matching.

\begin{figure}[t]
\centering
\includegraphics[width=\linewidth]{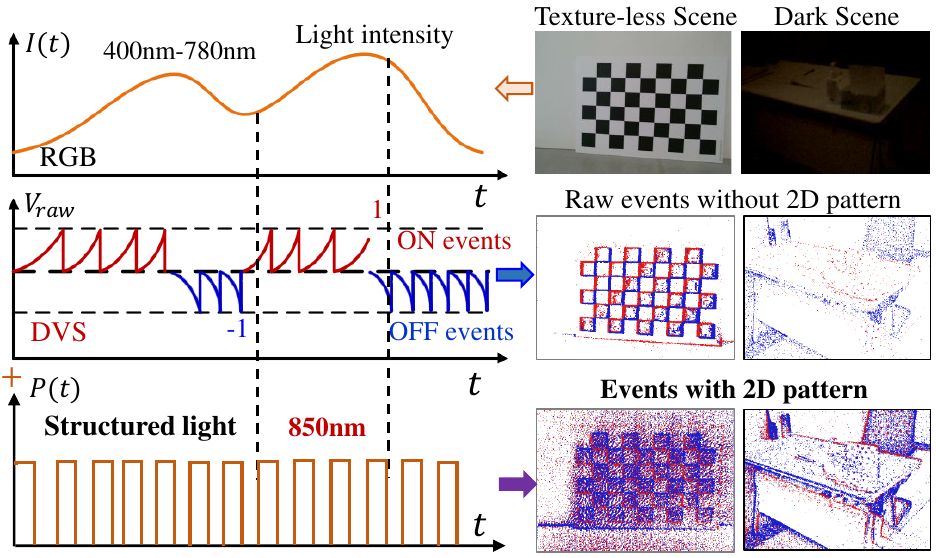}
\caption{Comparison with the event camera using or without structured light. Our solution enables event cameras to generate events even in static scenes or low-light conditions.}
\label{fig:light_curve}
\vspace{-0.30cm}
\end{figure}

\subsection{Event Camera Sampling Principle} \label{subsec:work_principle}
Event cameras~\cite{gallego2020event}, namely dynamic vision sensor (DVS)~\cite{lichtsteiner2008128}, respond to light changes with a continuous event stream. More specifically, each event $e_{n}$ is represented as a tuple $\langle x_{n}, y_{n}, t_{n}, p_{n} \rangle$ in address event representation (AER)~\cite{gallego2020event}, comprising spatial coordinates $\langle x_{n}, y_{n} \rangle$, a timestamp $t_{n}$, and the binary polarity $p_{n}$. Intuitively, these events form sparse asynchronous points in the spatiotemporal domain as:
\begin{eqnarray}
S\left(x,y,t\right)=\left\lbrace  p_{n}\delta\left(x-x_{n},y-y_{n},t-t_{n}\right)    \right\rbrace _{n=1}^{N_{e}}.
\end{eqnarray}
where the polarity $p_{n} \in \left\lbrace 1, -1 \right\rbrace $ denotes whether the brightness is increasing or decreasing, $N_{e}$ is the number of events in a spatiotemporal window, and $\delta\left(\cdot\right)$ refers to the Dirac delta function, with $\delta\left(t\right)=0, \forall$$t\neq0$ and $\int$$\delta\left(t\right)dt=1$.

This novel sensing mechanism enables event cameras to perform reliable depth estimation in high-speed motion scenarios~\cite{gallego2020event, ghosh2024event}. However, most existing event-based stereo vision systems belong to passive vision, typically detecting changes in light intensity within the 400–780 nm spectrum. As a result, passive stereo vision struggles in texture-less or low-light scenes. Notably, the sensor chip of event cameras (e.g., DAVIS346~\cite{taverni2018front}) is sensitive to a broader spectrum ranging from 300 to 1000 nm. To address this limitation, this study presents an active stereo system that combines binocular event cameras with an 850 nm infrared pattern projector. As shown in Fig.~\ref{fig:light_curve}, the proposed system actively generates events by modulating the laser’s frequency or intensity, enabling event generation even in static scenes with constant illumination or in extremely low-light conditions.

\begin{figure}[t]
\centering
\includegraphics[width=\linewidth]{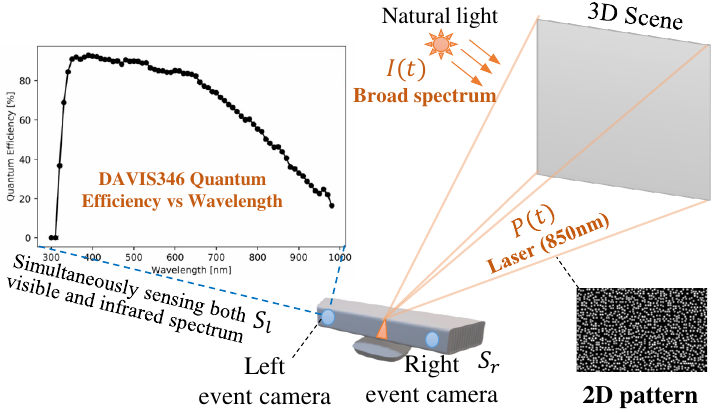}
\caption{Sensing mechanism of active event-based stereo vision. Projecting a 2D pattern from an infrared projector overcomes some challenges in texture-less regions or extremely low-light scenes, improved stereo matching performance.}
\label{fig:active_event_mechanism}
\vspace{-0.30cm}
\end{figure}

\subsection{Active Event-based Stereo Matching} \label{subsec:problem_definition}
This work aims to achieve high-speed depth sensing in challenging scenes by computing the disparity map $d_{e}$ from the stereo event streams $S_{l}$ and $S_{r}$. As illustrated in Fig.~\ref{fig:active_event_mechanism}, the generation of the event stream is mainly influenced by natural light, the 850 nm 2D laser pattern, and sensor noise, which can be mathematically formulated as:
\begin{align}
\begin{split}
  S_{l}= \mathcal{G}\left (W_{l} \cdot [\mathit{ I}\left ( t \right ) + \mathit{P}\left ( t \right )  ] +\mathcal{N} \left ( t \right )  \right ),\\
  S_{r}= \mathcal{G}\left (W_{r} \cdot [\mathit{I}\left ( t \right ) + \mathit{P}\left ( t \right )  ] +\mathcal{N} \left ( t \right )  \right ),
\end{split}
\end{align}
where $\mathcal{G}$ refers to the event generation process of event cameras. $W_{l}$ and $W_{r}$ denote the left and right warping operations via projecting a real 3D scene to each 2D camera plane. $\mathcal{N}\left ( t \right )$ is the event camera noise. $\mathit{I}(t)$ and $\mathit{P}(t)$ are the scene light intensity and the infrared laser intensity at time $t$. Consequently, our solution enables the generation of dynamic events in static scenes by modulating the intensity or frequency of the laser $\mathit{P}(t)$.

For learning-based stereo matching with asynchronous events, continuous event streams are typically divided into temporal bins. For example, the left event stream $S_{l}$ can be partitioned into $K$ temporal bins as $S_{l} = \left\lbrace S_{l,1},...,S_{l,K} \right \rbrace$. Given a timestamp $t$, the disparity map $D_{t}$ can be estimated using multiple left temporal bins $\left\lbrace S_{l, t-k}, ..., S_{l,t} \right \rbrace$, $k \in [1,K]$ and corresponding right temporal bins $\left\lbrace S_{r, t-k}, ..., S_{r,t} \right \rbrace$. Thus, this process can be formulated as follows:
\begin{eqnarray}
D_{t} = \mathcal{M}_{d}  ( \left\lbrace S_{l, t-k}, ..., S_{l,t} \right \rbrace, \left\lbrace S_{r, t-k}, ..., S_{r,t} \right \rbrace, \theta ),
\end{eqnarray}
where $\mathcal{M}_{d}$ refers to the proposed active event-based stereo matching model that effectively leverages rich temporal cues from $k+1$ adjacent stereo event bin pairs. The parameter $k$ controls the temporal bins for information aggregation, balancing the trade-off between accuracy and speed. $\theta$ denotes the optimized parameters of $\mathcal{M}_{d}$.

Given the ground truth $\bar{D}=\left\lbrace \bar{D}_{1}, ..., \bar{D}_{K}\right\rbrace $, we aim to make the predicted disparity maps $D=\left\lbrace D_{1}, ..., D_{K}\right\rbrace $ of the optimized model to fit $\bar{D}$ as closely as possible. This can be formulated as the following minimization problem:
\begin{equation}
\hat{\theta } = \mathop{\arg \min}_{\theta} \frac{1}{K} \sum\limits_{t=1}^{K} \mathcal{L_{M}} \left ( (D_{t}, \bar{D}_{t} \right )+\lambda \Phi \left ( \theta  \right ),
\end{equation}
where $\mathcal{L_{M}} \left ( D_{t}, \bar{D}_{t} \right )$ is the loss function between the predicted disparity $D_{t}$ and the ground truth $\bar{D}_{t}$. $\Phi \left ( \theta  \right )$ is the regularization term, and $\lambda$ is the trade-off parameter.

Note that, this novel active event-based stereo vision system operates fundamentally differently from conventional frame-based stereo cameras. Our solution has three unique properties: (i) \emph{It estimates depth accurately using temporal cues from stereo event streams instead of a single image pair}; (ii) \emph{Event streams with high temporal resolution enable on-demand continuous depth sensing, overcoming the fixed inference rates of conventional frame-based systems}; (iii) \emph{Structured light integration enables robust performance in static or low-light scenes where passive stereo vision systems fail}. 

\begin{figure*}[t]
\centering
\includegraphics[width=\linewidth]{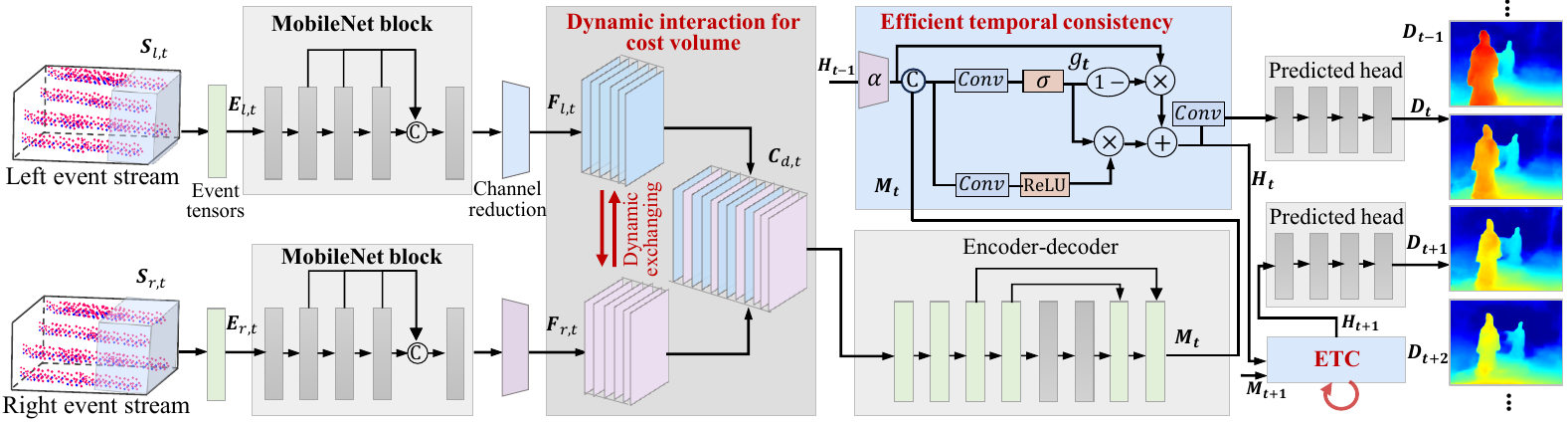}
\caption{Overview of \emph{active event-based stereo matching neural network (ActiveEventNet+)}. We then incorporate MobileNet blocks for feature extraction. Next, we design a novel cost volume with dynamic stereo event interactions, along with an efficient temporal consistency (ETC) module to use temporal cues. Finally, an encoder-decoder processes the cost volume features, which are refined by the temporal consistency module and decoded into dense disparity maps by a prediction head.}
\vspace{-0.20cm}
\label{fig:framework}
\end{figure*}

\section{Methodology} \label{sec:method}
This section begins with an overview of our framework (Section~\ref{subsec:network_overview}), followed by the incorporation of lightweight MobileNet blocks for stereo matching (Section~\ref{subsec:mobilenet}). We then detail the novel cost volume (Section~\ref{subsec:dynamic_interaction}) and the efficient temporal consistency architecture (Section~\ref{subsec:temporal_consitency}). Finally, we briefly describe the loss function (Section~\ref{subsec:loss_function}).

\subsection{Network Overview} \label{subsec:network_overview}
This work aims to design a novel lightweight yet effective active event-based stereo matching neural network, termed \emph{\textbf{ActiveEventNet+}}, which achieves ultra-high-speed dense depth sensing via integrating binocular event cameras and infrared structured light. As depicted in Fig.~\ref{fig:framework}, our framework mainly consists of five modules: event representation, feature extraction, \emph{\textbf{dynamic interaction for cost volume}}, \emph{\textbf{efficient temporal consistency}}, and an encoder-decoder. More specifically, the continuous event stream is first segmented into temporal event bins, each bin is converted into a 2D image-like representation (e.g., event images~\cite{maqueda2018event} or voxel grids~\cite{zhu2019unsupervised}) to optimize the trade-off between accuracy and speed. In the feature extraction stage, we incorporate lightweight MobileNet blocks~\cite{sandler2018mobilenetv2} with corresponding standard convolutions. The event embeddings from stereo pairs are fed into the feature extraction backbone and a channel reduction module to obtain compact yet powerful features. In particular, our ActiveEventNet+ contains two core innovations: (i) A dynamic 3D cost volume module, which enhances the estimated accuracy by dynamically exchanging feature channels and concatenating interactive stereo features; (ii) An efficient temporal consistency architecture, which exploits rich temporal cues from adjacent temporal event bins to achieve a balance between accuracy and speed. Finally, an encoder-decoder module processes the cost volume features and passes them to the temporal consistency component. The predicted head then refines the spatiotemporal features to generate dense disparity maps.

\subsection{Raising MobileNet for Event-based Stereo} \label{subsec:mobilenet}
To balance fidelity and inference speed, we utilize the lightweight MobileNet blocks~\cite{sandler2018mobilenetv2, shamsafar2022mobilestereonet} as a replacement for standard convolutional operations in active event-based stereo matching. As a pioneering work, MobileNet v1 introduces depthwise convolutions followed by pointwise convolutions, which dramatically reduce computational cost with minimal accuracy degradation compared to standard convolutions. Furthermore, MobileNet v2 improves efficiency by introducing linear bottlenecks and inverted residuals, enabling memory-efficient inference with low-latency. By leveraging these design principles, we aim to enhance the computational efficiency of event-based stereo matching while minimizing performance degradation. We formulate the output $F_{\textsl{res}}$ of an inverted residual block as follows:
\begin{align}
    \begin{split}
    & F_{1} = \sum_{c}^{tC} k_{p}(x,y,c) \cdot F(x, y, c), \\
    & F_{2} = \sum_{x, y} k_{d}(x, y, tc) \cdot F_{1}(x-1, y-1, tc), \\
    & F_{3} = \sum_{c} k_{p}(x,y,c) \cdot F_{2}(x, y, c), \\
    & F_{\textsl{res}} = F + F_{3}, 
    \end{split}
\end{align}
where $k_{d}(x,y,c)$ denotes the depthwise convolution kernel at position $(x,y,c)$ of the input feature map $F$ with channel number $C$. $k_{p}$ is the 1$\times$1 pointwise convolution kernel. $F_{1}$, $F_{2}$, and $F_{3}$ represent intermediate feature maps within the inverted residual block. The channel dimension $c$ is expanded by an expansion factor $t$ during the pointwise and depthwise convolution operations.

Inspired by MobileNet's computational efficiency, we integrate MobileNet v2 blocks into ActiveEventNet+, replacing standard convolutions in both the feature extraction and encoder-decoder modules. This design brings three key benefits: (i) Depthwise separable convolutions reduce computational cost by decoupling spatial and channel-wise processing; (ii) The linear bottleneck structure expands and compresses channel dimensions, preserving representational capacity while minimizing parameters; and (iii) Dynamic expansion ratios adapt to event sparsity patterns, focusing computation only where events occur. Together, these innovations enable low-latency processing of stereo event streams without compromising model accuracy.

\subsection{Dynamic Interaction for Cost Volume} \label{subsec:dynamic_interaction}

Cost volume~\cite{laga2020survey} is a fundamental component in the stereo matching pipeline, representing the matching cost between pixels at different disparities. A 3D cost volume $C_{d}$ can be constructed by computing the dissimilarity between the left feature map $F_{l}$ and the right feature map $F_{r}$ as follows:
\begin{eqnarray}
C_{d}(x, y, c)= \mathcal{M}_{c} (F_{l}(x, y, c),  F_{r}(x-d, y, c) ),
\end{eqnarray}
where $d$ denotes the disparity between pixels in a stereo pair, and $\mathcal{M}_{c}$ is the similarity measurement function. In learning-based models,  $\mathcal{M}_{d}$ typically uses concatenation or correlation operations~\cite{mostafavi2021event, nam2022stereo} between two feature maps.

This work aims at designing a lightweight yet powerful 3D cost volume that dynamically exchanges stereo channels and concatenates interacted features, achieving high performance with efficient computation. First, it employs context-aware dynamic gating, which computes channel-wise attention weights by jointly analyzing stereo features. This operation can be formally described as follows:
\begin{eqnarray}
G = \sigma \left(W_{g} * \text{AvgPool} \left( \text{Concat}(F_{l}, F_{r} ) \right) \right),
\end{eqnarray}
where $\text{Concat}(\cdot)$ is the concatenation operation for stereo features, $*$ represents the convolution, and $\sigma$ is the sigmoid function. $\text{AvgPool}(\cdot)$ refers to the global average pooling operation. $W_{g}$ is the weight of the 1$\times$1 convolution.

Then, feature adaptive weighting performs soft feature selection via channel-wise modulation. The stereo features are reweighted using complementary gating coefficients as:
\begin{align}
    \begin{split}
    & \tilde{F}_{l} = F_{l} \odot G, \\
    & \tilde{F}_{r} = F_{r} \odot (1 - G),
    \end{split}
\end{align}
where $\odot$ refers to channel-wise multiplication. This operation preserves discriminative features from each view while automatically suppressing unreliable regions.

Finally, the processed stereo features $F_{l}$ and $F_{r}$ are merged into the final cost volume $C_{d}$ via a dynamic channel interleaving operation. This strategy yields a compact yet expressive cost volume that maintains the geometric consistency critical for accurate disparity estimation with efficient computation. More specifically, each output channel $c$ is assigned from either the $i$-th channel of the left feature map $F_{l}$ or the right feature map $F_{r}$. The channel interleaving operation can be formally expressed as follows:
\begin{eqnarray}
C_{d}(x,y,c) = \sum_{i=1}^{C} \delta_{c,2i} \tilde{F}_{l}(x, y, i) + \delta_{c,2i+1} \tilde{F}_{r}(x, y, i) ,
\end{eqnarray}
where $i \in [1, C]$ denotes the input feature channel index, $c \in [1, 2C]$ is the output channel index of the cost volume, and $\delta_{c,2i}$ is the Kronecker delta function, which acts as a binary selector equal to 1 when $c = 2i$ and 0 otherwise.

\subsection{Efficient Temporal Consistency} \label{subsec:temporal_consitency}

Event streams offer rich temporal information, but existing event-based stereo matching models~\cite{ghosh2024event} often process each event bin independently using feed-forward architectures, discarding valuable temporal cues. While leveraging temporal cues may enhance depth accuracy and consistency, it typically incurs computational overhead. To balance this trade-off between accuracy and efficiency, we present an efficient temporal consistency (ETC) module for active event-based stereo matching, which refines depth estimates across event bins while minimizing redundant computation.

The ETC module adopts a recurrent architecture to process multiple adjacent event bins (see Fig.~\ref{fig:framework}). It is integrated after the encoder-decoder in our ActiveEventNet+. For each event bin, the ETC module takes as input the current feature map $M_{t}$ and the memory state $H_{t-1}$ propagated from previous bins. It then outputs a refined spatiotemporal feature map $M_{t}^{+}$ along with an updated memory state $H_{t}$. We can formulate the recurrent architecture $\mathcal{R}$ as follows:
\begin{eqnarray}
    (M_{t}^{+}, H_{t}) = \mathcal{R}(M_{t}, H_{t-1}).
\end{eqnarray}

To be specific, the ETC module incorporates three key optimization strategies for efficient processing: (i) reducing the number of hidden units, thereby decreasing both the parameter count and computational load; (ii) simplifying the activation function by replacing the tanh function with ReLU, eliminating costly exponential operations and improving runtime efficiency; and (iii) sharing gating parameters by merging the GRU’s reset and update gates into a single unified gate, computed through a linear layer to minimize parameter redundancy. Thus, the ETC module can be rewritten as follows:
\begin{align}
    \begin{split}
    & g_{t} = \sigma \left( W_{g} \cdot \text{Concat}(M_{t}, H_{t-1}) + b_{g} \right), \\
    & \tilde{H}_{t} = \text{ReLU} \left( W_{c} \cdot \text{Concat} (M_{t}, H_{t-1}) + b_{c} \right), \\
    & H_{t} = (1 - g_{t}) \odot H_{t-1} + g_{t} \odot \tilde{H}_{t}, \\
    & M_{t}^{+} = W_{o} \cdot H_{t} + b_o, 
    \end{split}
\end{align}
Where $M_t \in \mathbb{R}^C$ denotes the input feature vector at time step $t$, while $H_{t-1}, H_t, \tilde{H}_t \in \mathbb{R}^{H'}$ represent the previous hidden state, current hidden state, and candidate hidden state. The hidden dimension is reduced to $H'=\lfloor \alpha \cdot H \rfloor$, where $\alpha \in (0, 1)$ is a tunable compression ratio that balances model compactness and expressive capacity. The unified gate vector $g_t \in \mathbb{R}^{H'}$ is computed via a sigmoid function, based on the concatenated input and hidden features $[M_t, H_{t-1}] \in \mathbb{R}^{C + H'}$. The weight matrices $W_g, W_c \in \mathbb{R}^{H' \times (C + H')}$ and $W_o \in \mathbb{R}^{C \times H'}$, along with the bias terms $b_g, b_c \in \mathbb{R}^{H'}$ and $b_o \in \mathbb{R}^C$, are all learnable parameters. The functions $\sigma(\cdot)$ and $\text{ReLU}(\cdot)$ denote the sigmoid and rectified linear unit activations, respectively, and $\odot$ indicates element-wise multiplication. The final output $M_t^{+} \in \mathbb{R}^C$ is the temporally aggregated representation produced at each time step.

\subsection{Loss Functions} \label{subsec:loss_function}
Our ActiveEventNet+ adopts an end-to-end training strategy with full ground truth supervision across all predicted stages. The network produces $N$ progressive disparity refinement stages, comprising intermediate outputs and one final prediction. The total supervision loss is defined as:
\begin{eqnarray}
    \mathcal{L_{M}} = \sum_{n=1}^{N} W_{n} \cdot \mathcal{L}_{\text{smooth}}(D_{t}^{n}, \bar{D}_{t}),
\end{eqnarray}
Where $\mathcal{L}_{\text{smooth}}$ is the smooth $L_{1}$ loss. $W_{n}$ is the weight that increases to give higher weights for later predictions. Supervision is applied to all $N$ predicted stages during training, only the final-stage prediction is used during inference.

\begin{table*}[t]
\centering
\caption{Performance evaluation of ActiveEventNet+ on the real-world Active-Event-Stereo dataset. Notably, our event-based stereo vision solution with structured light outperforms frame-based stereo methods, particularly in low-light scenes.}
\renewcommand{\arraystretch}{1.05}
\setlength{\tabcolsep}{0.95 mm}{
\begin{tabular}{ll cccccc| cccccc}
\toprule
\multirow{2}*{Scenario}  & \multirow{2}*{Sequence} & \multicolumn{6}{c|}{RGB frames} & \multicolumn{6}{c}{Events} \\ \cline{3-8} \cline{9-14} & &  EPE$\downarrow$  & RMSE$\downarrow$  & D1-all$\downarrow$   & $>$1px$\downarrow$  & $>$2px$\downarrow$   & $>$3px$\downarrow$   & EPE$\downarrow$  & RMSE$\downarrow$  & D1-all$\downarrow$   & $>$1px$\downarrow$  & $>$2px$\downarrow$   & $>$3px$\downarrow$  \\ \hline
\multirow{5}{*}{Normal}
& 05\_indoor\_boxes &  1.752 & 7.962 & 0.089 & 0.244 & 0.126 & 0.088 & 1.793 & 7.955 & 0.068 & 0.299 & 0.119 & 0.068 \\
& 10\_indoor\_boxes & 2.534 & 8.916 & 0.175 & 0.534 & 0.302 & 0.175 & 2.492 & 8.848 & 0.158 & 0.580 & 0.301 & 0.157  \\
& 26\_indoor\_checkerboard & 1.908 & 7.875	& 0.099	& 0.363	& 0.159	& 0.099 & 1.722 & 7.725 & 0.058 & 0.289 & 0.102 & 0.063 \\ 
& 70\_outdoor\_car & 0.804 & 3.469 & 0.015 & 0.223 & 0.048 & 0.015 & 1.204 & 3.573 & 0.032 & 0.444 & 0.141 & 0.031 \\
& 80\_outdoor\_deer & 2.020 & 8.123 & 0.089 & 0.451 & 0.182 & 0.088 & 2.127 & 8.074 & 0.069 & 0.576 & 0.203 & 0.069 \\
\hline
\multirow{7}{*}{Low-light}
& 15\_indoor\_office\_desk\_dark & 2.901 & 11.787 & 0.124 & 0.383 & 0.184 & 0.123 & 3.031 & 12.004 & 0.141 & 0.392 & 0.209 & 0.140 \\
& 17\_indoor\_office\_desk\_dark & 3.342 & 11.670 & 0.210 & 0.591 & 0.355 & 0.210 & 2.753 & 11.699 & 0.099 & 0.354 & 0.166 & 0.098  \\
& 22\_indoor\_printer\_dark & 1.970 & 7.600 & 0.074 & 0.488 & 0.187 & 0.073 & 1.421 & 7.503 & 0.046 & 0.181 & 0.071 & 0.046 \\
& 27\_indoor\_checkerboard\_dark & 1.833 & 7.685 & 0.092 & 0.342 & 0.199 & 0.092 & 1.737 & 7.747  & 0.059 & 0.426 & 0.145 & 0.059 \\
& 32\_indoor\_conference\_desk\_dark & 2.624 & 7.091 & 0.253 & 0.653 & 0.413 & 0.253 & 1.450 & 6.655 & 0.051 & 0.305 & 0.121 & 0.050 \\
& 58\_indoor\_washroom\_dark & 1.651 & 5.526 & 0.072 & 0.529 & 0.208 & 0.071 & 1.072 & 5.356 & 0.029 & 0.222 & 0.062 & 0.029 \\
& 64\_outdoor\_car\_dark & 1.346 & 3.828 & 0.037 & 0.434 & 0.224 & 0.036 & 0.961 & 3.618 & 0.013 & 0.321 & 0.053 & 0.012 \\ 
\hline
\textbf{All} & \textbf{Average} & 2.128	& 7.850	& 0.121	& 0.444	& 0.221	& 0.120 & 1.849	& 7.787 & 0.073 & 0.351 & 0.141 &  0.072 \\
\bottomrule
\end{tabular}}
\label{table:sequence_results_real_dataset}
\end{table*}

\section{Experiments} \label{sec:experiment}
This section presents the experimental setup and implementation details (Section~\ref{subsec:experimental_setting}), followed by performance evaluations across diverse scenarios and comparisons with existing methods (Section~\ref{subsec:effective_test}). We then conduct ablation studies to analyze the impact of each module and parameter (Section~\ref{subsec:ablation_test}). Finally, we assess scalability to provide further insights into our ActiveEventNet+ (Section~\ref{subsec:scalability_test}).

\subsection{Experimental Settings} \label{subsec:experimental_setting}
In this subsection, we present the synthetic dataset, implementation details, and evaluation metrics as follows.

\textbf{\emph{Simulated Dataset}}. To provide a large amount of labor-saving yet high-quality synthetic data, we establish an active event-based stereo matching simulated dataset, named RealSense-Event-Sim. We first use an Intel RealSense D435 sensor to capture 119 infrared video sequences at a resolution of 640×480, covering variations in velocity, lighting conditions, and scene diversity. These videos are then converted into event streams using the V2E simulator~\cite{hu2021v2e}. The synthetic dataset provides stereo event streams along with 23.8k synchronized labels. We split the dataset into 16k samples for training, 3.8k for validation, and 4k for testing.

\textbf{\emph{Implementation Details}}. We utilize event images~\cite{gehrig2019end} as the event representation to balance accuracy and computational efficiency. The maximum disparity for stereo vision is set to 192 for stereo matching in all cases. The number of channels $C$ is set to 48 in implementing dynamic interaction for cost volume. Our ActiveEventNet+ incorporates an efficient temporal consistency module, using a temporal event bin size of 3 and a tunable compression ratio of 0.95 to balance accuracy and computational speed. For the training loss, we apply progressively increasing weights of [0.5, 0.5, 0.7, 1.0] across the four refinement stages, placing greater emphasis on later predictions~\cite{shamsafar2022mobilestereonet}. The proposed models and baselines are trained for 50 epochs using the Adam optimizer with a learning rate of $10^{-3}$. Inference time is measured on an NVIDIA RTX 3090 GPU. Final performance is reported using the best model selected on the validation set and evaluated on the test set.

\textbf{\emph{Evaluation Metrics}}. We evaluate accuracy using the mean average end-point error (EPE), root mean square error (RMSE), the percentage of pixels with a disparity error greater than 3 pixels or 5\% of the ground truth (D1-all), and the percentage of pixels with disparity errors exceeding 1, 2, and 3 pixels (i.e., $>$1px, $>$2px, and $>$3px). To assess computational efficiency, we report the number of model parameters (\#Params) and the average inference time.

\begin{table}[t]
\centering
\caption{Performance evaluation of ActiveEventNet+ on the simulated RealSense-Event-Sim dataset.}
\renewcommand{\arraystretch}{1.05}
\setlength{\tabcolsep}{1.00 mm}{
\begin{tabular}{ll ccc}
\toprule
Scenario  & Sequence &  EPE$\downarrow$  & RMSE$\downarrow$  & D1-all$\downarrow$  \\ \hline
\multirow{15}{*}{Normal}
& 002\_indoor\_boxes  & 0.746 & 1.266 & 0.023 \\
& 003\_indoor\_desk  & 1.156 & 1.844 & 0.079 \\ 
& 010\_indoor\_desk & 1.964 & 4.202 & 0.119  \\
& 042\_indoor\_car & 0.686 & 1.047 & 0.014  \\
& 047\_indoor\_office\_sofa & 1.606 & 4.326 & 0.062 \\
& 050\_indoor\_office\_cabinet & 1.005 & 1.670 & 0.054 \\
& 059\_indoor\_conference\_room & 0.786 & 1.162 & 0.023  \\
& 064\_indoor\_office\_checkerboard & 0.771 & 1.569 & 0.026  \\
& 073\_indoor\_floor & 1.803 & 3.392 & 0.132  \\
& 078\_indoor\_office\_room & 1.029 & 1.803 & 0.056  \\
& 090\_outdoor\_floor & 0.676 & 1.091 & 0.015 \\
& 097\_outdoor\_bridge & 0.507 & 0.979 & 0.016 \\
& 100\_outdoor\_people & 0.565 & 1.060 & 0.013 \\
& 106\_indoor\_checkerboard & 0.855 & 2.520 & 0.024 \\
& 107\_indoor\_checkerboard & 0.732 & 1.855 & 0.019 \\ \hline

\hline
\multirow{5}{*}{Low-light}
& 015\_indoor\_desk\_night & 0.971 & 1.526 & 0.049 \\
& 057\_indoor\_office\_night & 1.341 & 2.329 & 0.102  \\
& 108\_indoor\_checkerboard\_night & 0.715 & 1.039 & 0.017 \\ 
& 118\_outdoor\_wall\_night & 1.366 & 2.728 & 0.102 \\
& 119\_outdoor\_wall\_night & 1.770 & 3.358 & 0.130 \\

\hline
\textbf{All} & \textbf{Average} & 1.077 & 2.078 & 0.056	 \\
\bottomrule
\end{tabular}}
\label{table:sequence_results_simulated_dataset}
\end{table}

\begin{table*}[t]
\centering
\caption{Comparison with state-of-the-art methods on our real-world Active-Event-Stereo dataset. Our ActiveEventNet+, incorporating a lightweight yet effective recurrent architecture, outperforms nine state-of-the-art methods, including frame-based models, event-based models, and our baseline ActiveEventNet. ``+” indicates that a method leverages temporal cues.}
\renewcommand{\arraystretch}{1.05}
\setlength{\tabcolsep}{0.80 mm}{
\begin{tabular}{l cc cccccc cc}
\toprule
Method  & Event representation & Backbone & EPE $\downarrow$ & RMSE $\downarrow$ & D1-all $\downarrow$  & $>$1px $\downarrow$ & $>$2px $\downarrow$  & $>$3px $\downarrow$ & \# Params. (M) & Time (ms) \\ \hline
SGM~\cite{hirschmuller2007stereo} & Rec. images & No learning & 3.625 & 16.567 & 0.586 & 0.751 & 0.684 & 0.585 & - & 32.1 \\
PSMNet~\cite{chang2018pyramid} & Event images & 2D CNN & 2.894 & 11.756 & 0.204 & 0.603 & 0.352 & 0.204 & 5.22 & 15.6 \\
DeepPruner-Fast~\cite{duggal2019deeppruner} & Event images & 2D CNN & 2.514 & 8.758 &	0.113 &	0.545 &	0.253 &	0.112 &	7.39 & 39.4 \\
AANet~\cite{xu2020aanet} & Event images & 2D CNN & 2.317 & 8.123 &	0.106 &	0.525 &	0.287 &	0.106 &	3.68 &	16.7 \\
Unimatch~\cite{xu2023unifying} & Event images & Transformer & 1.902 & 7.759 & 0.068 & 0.361 & 0.116 & 0.066 & 4.70 & 87.6 \\
\hline
DDES~\cite{tulyakov2019learning} & Event embeddings & 2D CNN & 2.643 &	8.920 &	0.122 &	0.591 &	0.341 &	0.122 &	2.33 & 19.5 \\
StereoSpike~\cite{ranccon2022stereospike} & Event images & 2D SNN & 2.178 & 7.884 & 0.103 & 0.506 & 0.223 & 0.103 & 1.59 & 31.2 \\
SE-CFF~\cite{nam2022stereo} & Event stacks & 2D CNN & 2.015 & 7.824 & 0.089 & 0.425 & 0.182 & 0.088 & 27.00 & 24.9 \\
ActiveEventNet~\cite{li2025active} & Event images & MobileNet & 1.993 & 7.821 & 0.083 & 0.399 & 0.163 & 0.082 & 2.23 & \textbf{6.5} \\
\textbf{Our ActiveEventNet+} & Event images & MobileNet + ETC & \textbf{1.849}  & \textbf{7.787} & \textbf{0.073}  & \textbf{0.351} & \textbf{0.141} & \textbf{0.073} & 2.23 & 30.4 \\
\bottomrule
\end{tabular}}
\label{table:comparison_results_real_dataset}
\end{table*}

\begin{figure*}[t]
\centering
\includegraphics[width=\linewidth]{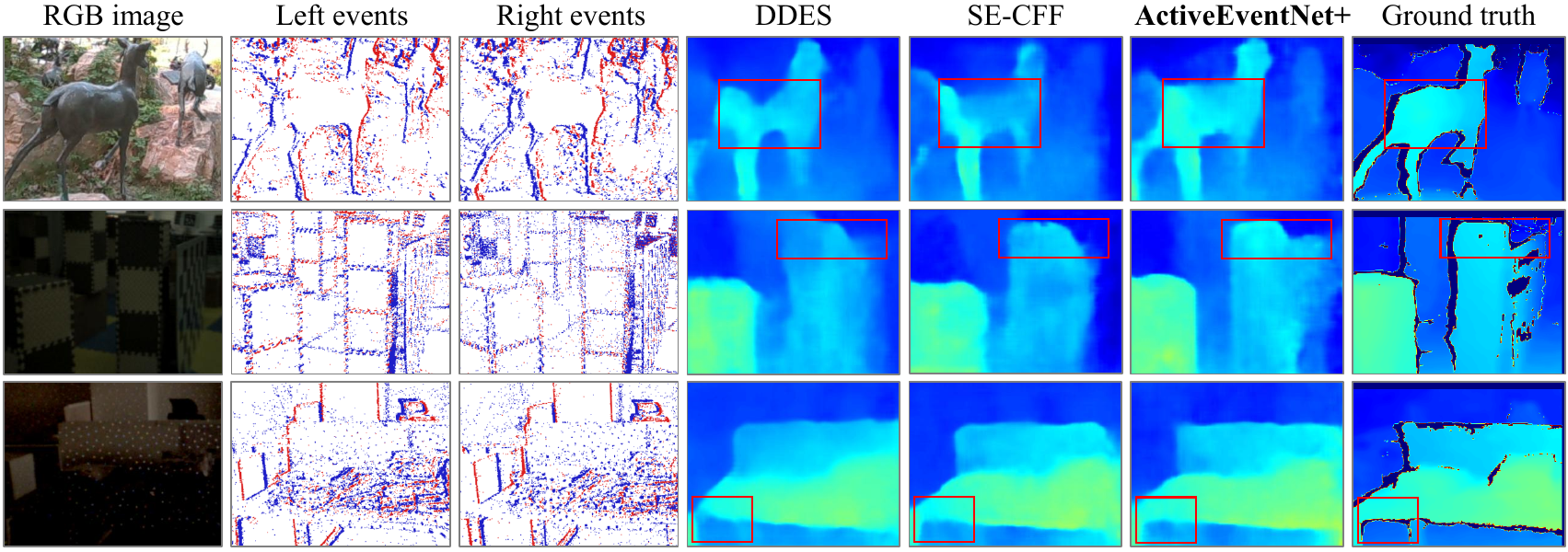}
\caption{Representative visualization results on our real-world Active-Event-Stereo dataset. The three rows show normal outdoor lighting, dark indoor environments, and low-light indoor scenes. ActiveEventNet+ outperforms two competing methods, delivering robust results across diverse lighting and scenes. Improvements are highlighted in red boxes.}
\label{fig:real_world_instances}
\end{figure*}

\begin{figure*}[t]
\centering
\includegraphics[width=\linewidth]{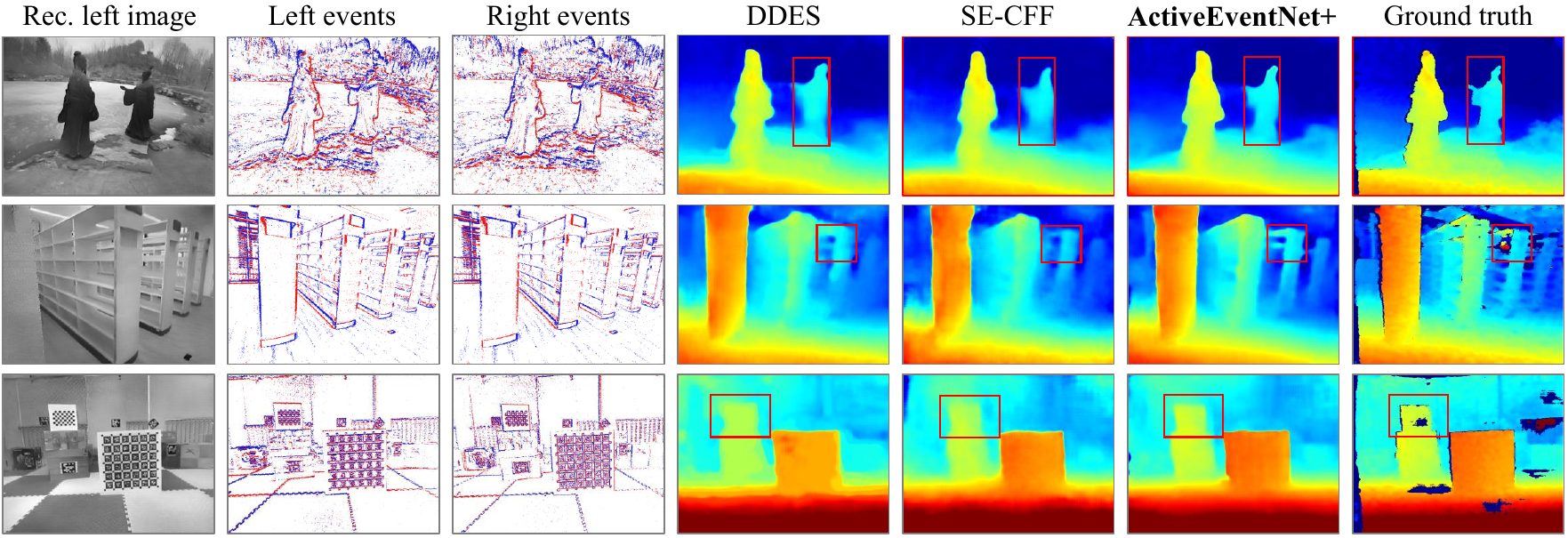}
\caption{Representative visualization results on our simulated RealSense-Event-Sim dataset. Our ActiveEventNet+ outperforms two state-of-the-art methods on both indoor and outdoor scenarios. Improvements are highlighted in red boxes.}
\label{fig:simulated_instances}
\vspace{-0.20cm}
\end{figure*}

\subsection{Effective Test} \label{subsec:effective_test}
This section verifies the effectiveness of our ActiveEventNet+ by conducting performance evaluations across various scenarios and comparing it with state-of-the-art methods.

\subsubsection{Performance Evaluation in Various Scenarios}
This subsection presents quantitative and visual results on both real-world and simulated datasets to provide a comprehensive evaluation across diverse scenarios.

\textbf{\emph{Evaluation on Real-world Dataset}}. To compare our solution with conventional frame-based stereo, we report quantitative results of ActiveEventNet+ for each sequence using both RGB frames and event data with structured light on the real-world Active-Event-Stereo dataset (see Table~\ref{table:sequence_results_real_dataset}). Our active event-based stereo vision solution is capable of producing high-quality dense disparity maps across both indoor and outdoor scenes under varying lighting conditions. It consistently outperforms passive stereo vision using RGB frames, especially in low-light scenarios. More specifically, ActiveEventNet+ significantly reduces errors across all evaluation metrics compared to passive stereo, with EPE, RMSE, and D1-all reduced by 0.279, 0.063, and 0.048, respectively. To our surprise, our solution using sparse stereo event streams outperforms dense RGB images in most sequences, even under normal light conditions. This performance gain can be attributed to the integration of structured light with event-based vision, which enhances scene texture and further improves disparity estimation accuracy.

\textbf{\emph{Evaluation on Simulated Dataset}}. To provide a detailed evaluation of ActiveEventNet+ on the simulated RealSense-Event-Sim dataset, we report the quantitative results for each synthetic sequence in Table~\ref{table:sequence_results_simulated_dataset}. The test set consists of 15 sequences under normal lighting and 6 sequences under low-light conditions. Note that, our ActiveEventNet+ obtains a satisfactory performance, achieving an average EPE of 1.077, RMSE of 2.078, and a D1-all of 0.056. The results show that our active event-based stereo vision scheme is capable of achieving high-quality depth sensing in both indoor and outdoor scenes across diverse lighting conditions.

\subsubsection{Comparison with State-of-the-Art Methods}
We investigate why and how ActiveEventNet+ outperforms other competitors from the following two perspectives.

\textbf{\emph{Quantitative Evaluation}}. To ensure a fair comparison with stereo matching methods, we first convert the event streams into videos using E2VID~\cite{rebecq2019events}, and then use the reconstructed grayscale images as input to the classical SGM algorithm~\cite{hirschmuller2007stereo}. In addition, we compare our ActiveEventNet+ with five frame-based stereo matching networks (i.e., PSMNet~\cite{chang2018pyramid}, DeepPruner-Fast~\cite{duggal2019deeppruner}, AANet~\cite{xu2020aanet} and Unimatch~\cite{xu2023unifying}), three open-source event-based stereo matching models (i.e., DDES~\cite{tulyakov2019learning}, StereoSpike~\cite{ranccon2022stereospike}, and SE-CFF~\cite{nam2022stereo}), and our baseline ActiveEventNet~\cite{li2025active}. For \emph{the real-world dataset evaluation}, we compare our ActiveEventNet+ against the above methods on the real-world Active-Event-Stereo dataset in Table~\ref{table:comparison_results_real_dataset}. Note that, our baseline ActiveEventNet outperforms most state-of-the-art feed-forward methods while maintaining a smaller model size and faster inference speed. Our ActiveEventNet+ incorporates an efficient temporal consistency module, which significantly boosts performance but also increases computational cost. Compared to the top-performing frame-based stereo matching model Unimatch~\cite{xu2023unifying}, our ActiveEventNet+ achieves higher accuracy with a 2.9$\times$ speedup in inference time. For \emph{the simulated dataset evaluation}, we report the partial quantization metrics of our ActiveEventNet+ and other competitors on the synthetic RealSense-Event-Sim dataset in Table~\ref{table:comparison_results_simulated_dataset}. The results show that the trends observed on the simulated dataset align well with those from the real-world dataset. In short, our ActiveEventNet+, built on a lightweight yet powerful recurrent architecture, consistently outperforms nine state-of-the-art methods. This strong balance between accuracy and speed makes our ActiveEventNet+ well-suited for applications that demand high-precision depth sensing under challenging conditions.

\begin{table}[t]
\centering
\caption{Comparison with state-of-the-art stereo matching methods on our simulated RealSense-Event-Sim dataset.}
\renewcommand{\arraystretch}{1.05}
\setlength{\tabcolsep}{1.55 mm}{
\begin{tabular}{l cc cc}
\toprule
Method  & EPE $\downarrow$ & RMSE $\downarrow$ & D1-all $\downarrow$ & Runtime (ms) \\ \hline
SGM~\cite{hirschmuller2007stereo} & 4.298 & 8.326 & 0.681 & 43.2 \\
PSMNet~\cite{chang2018pyramid} & 2.786 & 5.583 & 0.211 & 20.6 \\
DeepPruner-Fast~\cite{duggal2019deeppruner} & 1.463 & 2.513 & 0.087 & 76.0 \\
AANet~\cite{xu2020aanet} &  1.397 &	2.463 &	0.077 & 28.5 \\
Unimatch~\cite{xu2023unifying} & 1.108 & 1.780 & 0.063 & 172.6 \\ 
\hline
DDES~\cite{tulyakov2019learning} &  1.696 &	3.241 &	0.125 & 36.4 \\
StereoSpike~\cite{ranccon2022stereospike} & 1.500 & 2.718 & 0.096 & 73.6 \\
SE-CFF~\cite{nam2022stereo} & 1.308 & 2.326 & 0.081 & 65.3 \\
ActiveEventNet &  1.223 &	2.320 & 0.070 & \textbf{21.9} \\
\textbf{Our ActiveEventNet+} & \textbf{1.077} & \textbf{2.078} & \textbf{0.056} & 97.3 \\
\bottomrule
\end{tabular}}
\label{table:comparison_results_simulated_dataset}
\end{table}

\textbf{\emph{Visualization Evaluation}}. To provide an intuitive qualitative evaluation, Figures~\ref{fig:real_world_instances} and \ref{fig:simulated_instances} present representative visualization results from the real-world Active-Event-Stereo dataset and the simulated RealSense-Event-Sim dataset, respectively. Note that, our ActiveEventNet+ consistently outperforms two state-of-the-art event-based stereo matching methods (i.e., DDES~\cite{tulyakov2019learning} and SE-CFF~\cite{nam2022stereo}). In dark indoor environments, where RGB images provide little to no discernible structure, our method effectively leverages stereo events to recover meaningful scene contours. In particular, both feed-forward models fail to capture edge details, whereas ActiveEventNet+ generates sharp and accurate disparity boundaries (see highlighted in red boxes). This improvement stems from the proposed lightweight yet efficient temporal consistency module in ActiveEventNet+, which exploits rich temporal cues from stereo event streams. Overall, our ActiveEventNet+ excels at producing high-quality disparity maps, even under varying lighting conditions in both indoor and outdoor environments.

\begin{table*}[t]
\centering
\caption{The contribution of each component to our ActiveEventNet+ on our real-world Active-Event-Stereo dataset and our simulated RealSense-Event-Sim dataset. All results are based on the baseline model, which uses standard convolutions, simple concatenation in the cost volume module, and does not incorporate the efficient temporal consistency (ETC) module.}
\renewcommand{\arraystretch}{1.15}
\setlength{\tabcolsep}{1.35 mm}{
\begin{tabular}{l ccc cccc| cccc}
\toprule
\multirow{2}*{Method}  & \multirow{2}*{MobileNet}  & \multirow{2}*{Cost volume}  & \multirow{2}*{ETC} & \multicolumn{4}{c|}{Active-Event-Stereo dataset} & \multicolumn{4}{c}{RealSense-Event-Sim dataset} \\ \cline{5-8} \cline{9-12} & & & &  EPE$\downarrow$  & RMSE$\downarrow$  & D1-all$\downarrow$   &  Runtime (ms)    & EPE$\downarrow$  & RMSE$\downarrow$  & D1-all$\downarrow$   &  Runtime (ms)    \\ \hline
Baseline & & & & 2.124 & 7.935 & 0.092 & 33.8 & 1.468 & 2.496 & 0.109 &  108.1 \\
(a) & \ding{51} & & & 2.210 & 7.952 & 0.095 & \textbf{6.2} & 1.516 & 2.972 & 0.117 & \textbf{20.2} \\
(b) & & \ding{51} & & 1.968 & 7.750 & 0.075 & 35.7 & 1.209 & 2.294 & 0.067 &  112.3 \\
(c) & \ding{51} & & \ding{51} & 2.015 & 7.824 & 0.088 & 29.0  & 1.153 & 2.185 & 0.058 & 95.1 \\
ActiveEventNet & \ding{51} & \ding{51} & & 1.993 & 7.821 & 0.083 & 6.5 & 1.223 & 2.320 & 0.070 & 21.9 \\
\textbf{Our ActiveEventNet+} & \ding{51} & \ding{51} & \ding{51} & \textbf{1.849} & \textbf{7.787} & \textbf{0.073} & 30.4 & \textbf{1.077} & \textbf{2.078} & \textbf{0.056} & 97.3 \\
\bottomrule
\end{tabular}}
\label{table:ablation_test}
\end{table*}

\begin{table}
\centering
\caption{Comparison of ActiveEventNet+ with event representations on the real-world Active-Event-Stereo dataset.}
\renewcommand{\arraystretch}{1.05}
\setlength{\tabcolsep}{1.00mm}{
\begin{tabular}{l cccc}
    \toprule
    Event representation & EPE $\downarrow$ & RMSE$\downarrow$ & D1-all $\downarrow$ & Runtime (ms) \\
    \hline
    Event images~\cite{gehrig2019end} & 1.849 & 7.787 & 0.073 & \textbf{30.4} \\
    Voxel grid~\cite{zhu2019unsupervised} & 1.827 & 7.800 & 0.075 & 33.7 \\
    Event embeddings~\cite{tulyakov2019learning} & 1.819 & 7.780 & 0.073  &  38.2 \\
    Reconstructed images~\cite{rebecq2019events} & \textbf{1.791} & \textbf{7.761} & \textbf{0.072} & 59.1 \\
    \bottomrule
\end{tabular}}
\label{table:event_representations}
\end{table}

\begin{table}[t]
\centering
\caption{The influence of MobileNet blocks in ActiveEventNet+ on our real-world Active-Event-Stereo dataset.}
\renewcommand{\arraystretch}{1.05}
\setlength{\tabcolsep}{0.60mm}{
\begin{tabular}{lc cccc}
    \toprule
    Feature extraction & Encoder-decoder  & EPE$\downarrow$ & RMSE$\downarrow$  & D1-all$\downarrow$ & Time (ms) \\
    \hline
    Standard Conv. & Standard Conv. & \textbf{1.752} & \textbf{7.756} & \textbf{0.070} &  156.9 \\
    Standard Conv. & MobileNet v2 & 1.819 & 7.759 & 0.072 & 98.8 \\
    MobileNet v2 & Standard Conv. & 1.825 & 7.770 & 0.072 & 61.6 \\
    MobileNet v2 & MobileNet v2 & 1.849 & 7.787 & 0.073 & \textbf{30.4} \\
    \bottomrule
\end{tabular}}
\label{table:mobilenet_blocks}
\end{table}

\subsection{Ablation Test} \label{subsec:ablation_test}
This section performs ablation studies to assess how design choices and parameters affect performance as follows.

\subsubsection{Contribution of Each Component}
To explore the impact of each component on final performance, the baseline model uses standard convolutions, simple concatenation in the cost volume module, and excludes the efficient temporal consistency (ETC) module. As illustrated in Table~\ref{table:ablation_test}, five event-based stereo matching methods, namely (a), (b), (c), ActiveEventNet, and our ActiveEventNet+, consistently indicate that leveraging MobileNet blocks boosts inference speed, while the introduction of dynamic interaction for cost volume and the efficient temporal consistency (ETC) module enhances accuracy on two datasets. More specifically, incorporating dynamic interaction for the cost volume results in EPE reductions of 0.156 and 0.259, RMSE decreases of 0.185 and 0.202, and D1-all declines of 0.017 and 0.042 across the two datasets. Compared to ActiveEventNet, our ActiveEventNet+ reduces the EPE from 1.993 to 1.849, the RMSE from 7.821 to 7.787, and the D1-all from 0.083 to 0.073 on the real-world Active-Event-Stereo dataset. Similarly, it lowers the EPE from 1.223 to 1.077, the RMSE from 2.320 to 2.078, and the D1-all from 0.070 to 0.056 on the simulated RealSense-Event-Sim dataset. To our surprise, replacing standard convolutions with MobileNet blocks results in nearly 5.5× and 5.4× increases in inference speed on the real-world and simulated datasets, respectively. Nevertheless, the efficient temporal consistency (ETC) module improves accuracy by leveraging rich temporal cues from stereo event streams, but it also increases computational complexity. We believe this accuracy-speed trade-off can be valuable in depth sensing scenarios where higher precision is required.

\subsubsection{Influence of Event Representation}
To assess the generality of our ActiveEventNet+ across different event representations, we compare four typical event representations (i.e., event images~\cite{gehrig2019end}, voxel grids~\cite{zhu2019unsupervised}, event embeddings~\cite{tulyakov2019learning}, and reconstructed images~\cite{rebecq2019events}) on the real-world Active-Event-Stereo dataset. As shown in Table~\ref{table:event_representations}, improved disparity estimation accuracy often comes at the cost of increased computational overhead. In this study, we choose to encode event temporal bins into event images to balance accuracy and inference speed. Notably, our ActiveEventNet+ provides a flexible interface compatible with various input event representations. We believe that a simple yet effective event representation enables our ActiveEventNet+ to better exploit stereo event streams, achieving high accuracy with low computational cost.

\subsubsection{Influence of MobileNet Block}
To evaluate the impact of MobileNet blocks in our ActiveEventNet+, we replace standard convolutions with MobileNet v2 blocks in both the feature extraction and encoder-decoder modules. As illustrated in Table~\ref{table:mobilenet_blocks}, incorporating MobileNet v2 blocks allows ActiveEventNet+ to achieve significantly faster inference speeds while maintaining comparable accuracy. More specifically, using MobileNet v2 blocks in the feature extraction and encoder-decoder modules results in only a slight drop in accuracy, while reducing inference time by approximately 2.5$\times$ and 1.6$\times$, respectively. Overall, the absolute increase in EPE is just 0.097, while the total inference time is reduced by nearly 5.2$\times$. The results demonstrate that using MobileNet v2 enables our ActiveEventNet+ to achieve an effective trade-off between accuracy and computational efficiency, making it well-suited for practical applications.

\begin{table}
\centering
\caption{Comparison of ActiveEventNet+ with various cost volume strategies on our Active-Event-Stereo dataset.}
\renewcommand{\arraystretch}{1.05}
\setlength{\tabcolsep}{1.30mm}{
\begin{tabular}{l ccccc}
    \toprule
    Aggregation operation & EPE $\downarrow$ & RMSE $\downarrow$ & D1-all $\downarrow$ & Runtime (ms) \\
    \hline
    Concatenation & 2.015 & 7.824 & 0.088  & \textbf{29.0} \\
    Correlation~\cite{nam2022stereo} & 1.939 & 7.813 & 0.079  & 29.5 \\
    Attention-based fusion & 1.879 & 7.782 &  0.078 & 29.8 \\
    \textbf{Dynamic interaction} & \textbf{1.849}  & \textbf{7.787} & \textbf{0.073} & 30.4 \\
    \bottomrule
\end{tabular}}
\label{table:cost_volumes}
\end{table}

\begin{table}
\centering
\caption{The influence of temporal modeling architectures in ActiveEventNet+ on our Active-Event-Stereo dataset.}
\renewcommand{\arraystretch}{1.05}
\setlength{\tabcolsep}{0.55mm}{
\begin{tabular}{l ccccc}
    \toprule
    Modeling architecture & EPE $\downarrow$ & RMSE $\downarrow$ & D1-all $\downarrow$ & Runtime (ms) \\
    \hline
    Our feed-forward baseline~\cite{li2025active} & 1.993 & 7.821 & 0.083 & \textbf{6.5} \\
    ConvLSTM~\cite{li2022asynchronous} & 1.904 & 7.830 & 0.089  & 43.3 \\
    ConvGRU~\cite{gehrig2021combining} & \textbf{1.843} & 7.803 & 0.080 & 38.9 \\
    \textbf{Our ETC} & 1.849  & \textbf{7.787} & \textbf{0.073} & 30.4 \\
    \bottomrule
\end{tabular}}
\label{table:temporal_modeling}
\end{table}

\subsubsection{Influence of Cost Volume Strategy}
To evaluate the effectiveness of various cost volume strategies in our ActiveEventNet+, we compare our approach with several typical aggregation methods on the Active-Event-Stereo dataset. We set the number of channels $C$ to 48 for all aggregation operations used to construct the cost volume. As depicted in Table~\ref{table:cost_volumes}, our dynamic interaction strategy achieves the best performance compared to three methods (i.e., concatenation, correlation~\cite{nam2022stereo}, and attention-based fusion) while maintaining comparable computational speed. This improvement is attributed to the ability of our dynamic interaction strategy to efficiently exchange stereo information and aggregate the interacted features. The results show that our method enables the construction of a lightweight yet powerful 3D cost volume, achieving high accuracy with efficient computation.

\begin{figure*}[t]
\centering
\includegraphics[width=\linewidth]{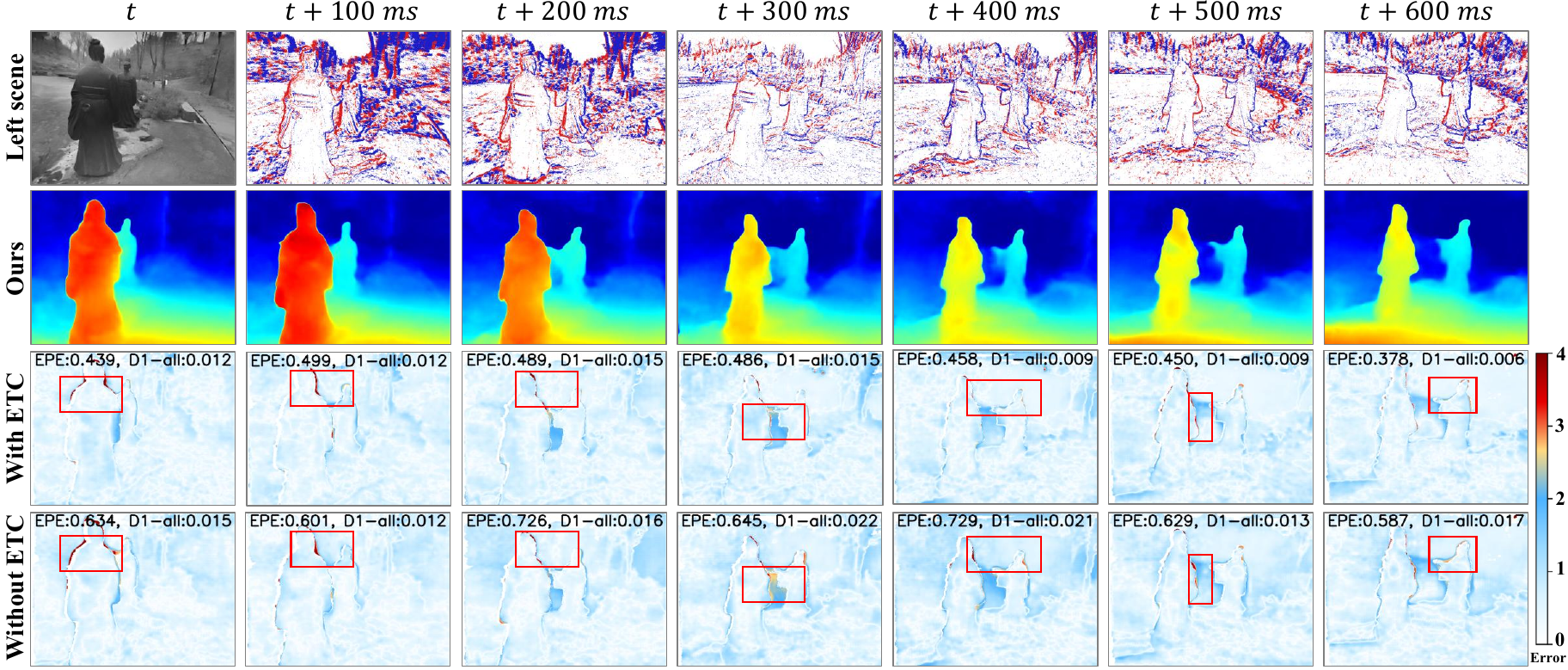}
\caption{Representative visualization results on continuous sequences from the RealSense-Event-Sim dataset. The four rows show the left event stream sequence, predicted disparity maps from our ActiveEventNet+, error maps with our efficient temporal consistency (ETC) module, and error maps without ETC. Note that, our ActiveEventNet+ using temporal cues outperforms the feed-forward baseline (i.e., ActiveEventNet~\cite{li2025active}), with improvements highlighted in red boxes.}
\label{fig:temporal_instances}
\vspace{-0.20cm}
\end{figure*}

\subsubsection{Influence of Temporal Modeling Architecture}
To evaluate the impact of different temporal modeling architectures, we assess the effectiveness of the proposed efficient temporal consistency (ETC) module by replacing it with typical recurrent temporal models (e.g., ConvLSTM~\cite{li2022asynchronous} and ConvGRU~\cite{gehrig2021combining}) in our ActiveEventNet+. As shown in Table~\ref{table:temporal_modeling}, all recurrent architectures outperform the feed-forward model (i.e., ActiveEventNet) without temporal consistency modeling. These results highlight the importance of a well-designed temporal consistency strategy for achieving optimal performance. Compared to the two alternatives, our ETC module achieves higher accuracy and lower inference time than ConvLSTM, and offers faster inference than ConvGRU while maintaining comparable accuracy.

To further enhance the interpretability of temporal modeling, we provide comparative visualizations to assess whether temporal cues are effectively exploited. As shown in Fig.~\ref{fig:temporal_instances}, we present four temporal sequences for each timestamp: the left event stream, predicted disparity maps from our ActiveEventNet+, corresponding error maps with the proposed efficient temporal consistency (ETC) module, and error maps without ETC. The feed-forward baseline (i.e., ActiveEventNet) exhibits progressively larger disparity errors over time, as reflected in its higher EPE and D1-all scores. In contrast, our ActiveEventNet+, incorporating the ETC module, produces disparity predictions with consistently lower errors across each timestamp, underscoring the impact of ETC on accuracy. These results confirm that the ETC module effectively models temporal dependencies in stereo event streams, enabling ActiveEventNet+ to achieve more accurate and temporally stable disparity estimation.

\begin{table}
\renewcommand{\arraystretch}{1.05}
\caption{The influence of temporal aggregation length on our Active-Event-Stereo dataset. The feed-forward baseline is conducted using an event bin for ActiveEventNet~\cite{li2025active}.}
\label{table:temporal_aggegation}
\setlength{\tabcolsep}{1.80mm}{
\begin{tabular}{c cccc}
    \hline
    Aggregation length & EPE $\downarrow$ &  RMSE $\downarrow$ & D1-all $\downarrow$ & Runtime ($ms$)  \\
    \hline
    1 & 1.995 & 7.821 & 0.083 & \textbf{6.5} \\
    2 & 1.890 & 7.812 & 0.082 & 20.3 \\
    3 & 1.849 & 7.787 & 0.073 & 30.4 \\
    4 & 1.829 & 7.792 & 0.072 & 53.7 \\
    6 & \textbf{1.782} & \textbf{7.761} & \textbf{0.071} & 76.1 \\
    \hline
\end{tabular}}
\end{table}

\subsubsection{Influence of Temporal Aggregation Length}
To assess the influence of temporal parameters in the ETC module, we conduct an ablation study to investigate how the temporal aggregation length affects performance on the Active‑Event‑Stereo dataset. As shown in Table~\ref{table:temporal_aggegation}, we evaluate ETC with aggregation lengths of 1, 2, 3, 4, and 6 event bins. Compared with the feed‑forward baseline, the corresponding EPE values are reduced by 0.105, 0.146, 0.166, and 0.213, respectively. In addition, both RMSE and D1‑all consistently decrease as the temporal aggregation length increases, suggesting that longer aggregation allows the ETC module to capture richer long‑range temporal cues from stereo event streams, thereby improving accuracy. However, this improvement comes at the cost of higher computational time. To balance accuracy and computational speed, we set the temporal aggregation length to 3 in our ETC module. This trade‑off is particularly well‑suited for some scenarios demanding higher accuracy.

\subsection{Scalability Test} \label{subsec:scalability_test}
This section investigates the role of structured light, evaluates the effectiveness of our solution in high‑speed scenes, and explores the feasibility of integrating frames and events.

\subsubsection{Analyzing the Role of Structured Light}

\begin{figure}[t]
\centering
\includegraphics[width=\linewidth]{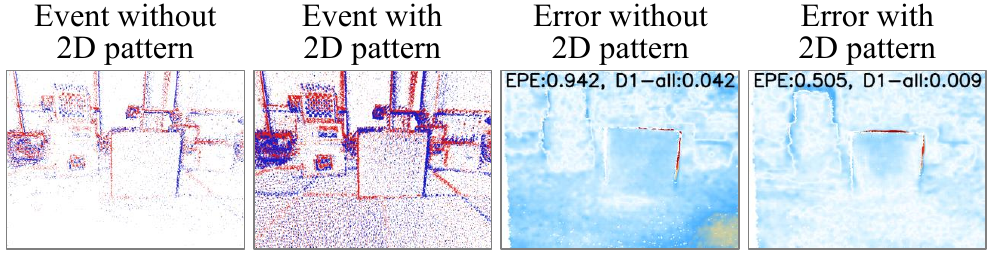}
\caption{Representative example with and without structured light in extremely slow‑motion scenes. Unlike passive stereo, our active stereo vision solution generates more events to enhance texture and achieves better performance.}
\label{fig:structured_light_result}
\vspace{-0.20cm}
\end{figure}

To evaluate the impact of structured light on event‑based stereo vision, we conduct a comparative experiment on a representative slow‑motion sequence (see Fig.~\ref{fig:structured_light_result}). Specifically, we fix the camera system and apply slight perturbations along a predefined trajectory, performing experiments both with and without an infrared filter placed in front of the event‑camera lens to control the presence of structured light. The quantitative results show that incorporating structured light significantly improves accuracy. In addition, we observe that passive stereo vision generates almost no events, whereas our prototype produces abundant dynamic events thanks to structured light. In other words, integrating structured light into binocular event‑camera systems can effectively overcome the limitations of passive event‑based stereo schemes and enhance texture representation in static scenes or low‑light environments.

\begin{figure}[t]
\centering
\includegraphics[width=\linewidth]{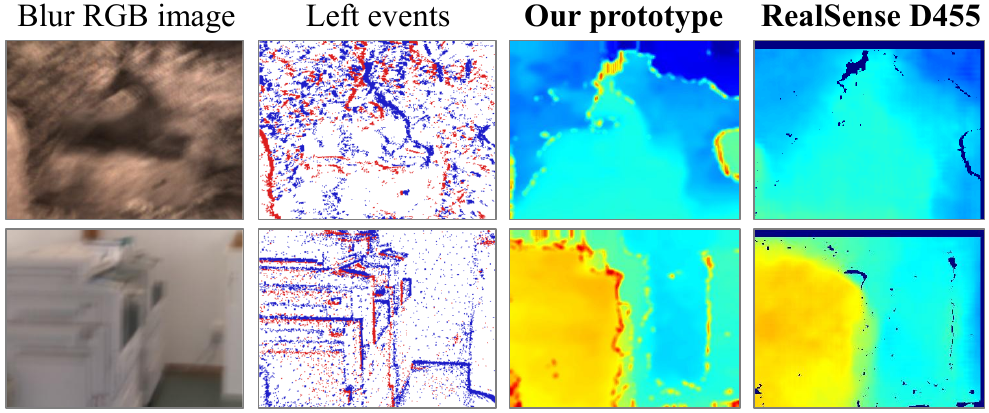}
\caption{Comparison with frame-based active stereo camera in high-speed motion scenes. Notably, our camera prototype outperforms Realsense D455 for high-speed depth sensing.}
\label{fig:high_speed_motion}
\vspace{-0.35cm}
\end{figure}

\begin{table}[t]
\centering
\renewcommand{\arraystretch}{1.05}
\caption{The impact of different modalities on the Active-Event-Stereo dataset. The joint integration of frames and events outperforms each single-modality method.}
\label{table:frames_events}
\setlength{\tabcolsep}{0.30mm}{
\begin{tabular}{cc c cccc}
    \toprule    
    \multicolumn{2}{c}{Input rate (Hz)} & \multirow{2}{*}{Modality} & \multirow{2}{*}{EPE $\downarrow$} & \multirow{2}{*}{RMSE $\downarrow$} & \multirow{2}{*}{D1-all $\downarrow$} & \multirow{2}{*}{Runtime (ms)} \\
    \cline{1-2} Frame & Label & & & & \\
    \hline
    0 & 15 & Events & 1.849 & 7.787 & 0.073 & 30.4 \\
    \hline
    \multirow{2}{*}{5} & \multirow{2}{*}{15} & Frames & 2.343 & 7.959 & 0.141 & 30.2 \\
    & & Events + Frames & 1.828 & 7.765 & 0.072 & 31.5 \\
    \hline
    \multirow{2}{*}{10} & \multirow{2}{*}{15} & Frames & 2.223 & 7.898 & 0.126 & 30.2 \\
    & & Events + Frames & 1.819 & 7.763 & 0.072 & 31.5 \\
    \hline
    \multirow{2}{*}{15} & \multirow{2}{*}{15} & Frames & 2.128	& 7.850	& 0.121 & 30.2 \\
    & & Events + Frames & \textbf{1.767} & \textbf{7.759} & \textbf{0.071} & 31.5 \\
    \bottomrule
\end{tabular}}
\vspace{-0.15cm}
\end{table}

\subsubsection{Verification in High-Speed Scenarios}
To validate our high-speed depth sensing solution, we compare our camera prototype with a conventional frame-based stereo system (Intel RealSense D455). As illustrated in Fig.~\ref{fig:high_speed_motion}, our method achieves higher disparity estimation accuracy than the RealSense D455 at 90 FPS in high-speed motion scenarios. Conventional RGB frames from the RealSense D455 suffer from motion blur, which hinders high-quality depth perception in such conditions. In contrast, our system leverages the high temporal resolution of event cameras to enable high-frequency depth sensing, while incorporating structured light to enhance texture and further improve accuracy. More precisely, the proposed lightweight ActiveEventNet efficiently processes stereo event stream pairs, achieving inference speeds of up to 150 FPS on an NVIDIA 3090 GPU. Moreover, the proposed ActiveEventNet+ offers a flexible trade-off between accuracy and speed by exploiting rich temporal cues, making it well-suited for specific scenarios that demand higher accuracy.

\begin{figure}[t]
\centering
\includegraphics[width=\linewidth]{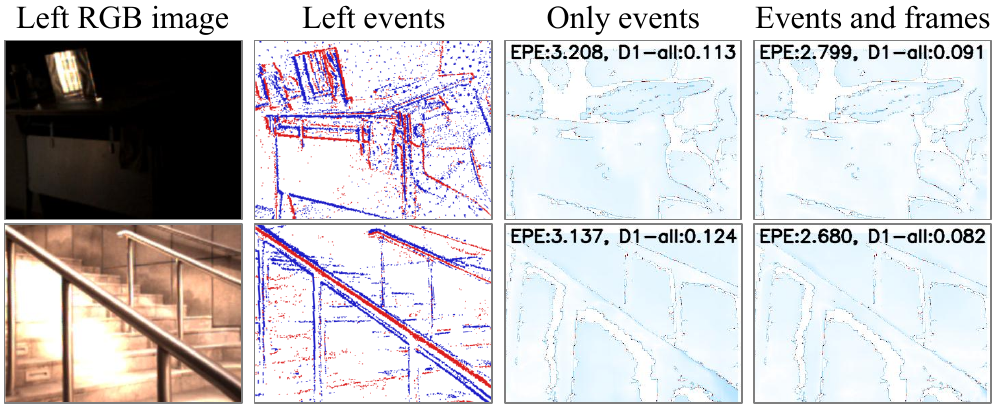}
\caption{Representative visualization results in indoor and outdoor scenarios. Incorporating frames improves the performance of the active event-based stereo matching task.}
\label{fig:modality_instances}
\vspace{-0.35cm}
\end{figure}

\subsubsection{Depth Sensing with Events and Frames}
To assess whether frames are necessary for active event-based stereo vision, we develop a joint framework that integrates frames and events for stereo matching on the Active-Event-Stereo dataset. Specifically, we utilize an asynchronous attention-based fusion module~\cite{li2023sodformer} to integrate the cost volumes from stereo event streams and frames in our ActiveEventNet+. To verify the effectiveness of the asynchronous fusion strategy, we vary only the frame rate while keeping the temporal event bins fixed at 15 Hz, with the output frequency remaining the same. The baseline refers to our ActiveEventNet+, which processes only the frame modality without using the fusion module. As shown in Table~\ref{table:frames_events}, combining the two modalities consistently reduces three evaluation metrics (i.e., EPE, RMSE, and D1-all) compared to using events or frames alone. Meanwhile, the joint framework achieves computational speed comparable to single-modality methods. Due to the limited measurement rate of the RealSense D455, the real-world dataset provides depth ground truth at only 15 Hz. Ideally, the inference frequency of the joint framework could reach the ultra-high temporal resolution of an event camera, depending on the application and available computational capacity.

Furthermore, Fig.~\ref{fig:modality_instances} presents representative indoor and outdoor results in challenging scenes. The visualizations clearly show that incorporating frames further improves event-based stereo matching, particularly in low-texture regions. This is because conventional frames provide valuable texture information for static or slow-motion scenes, while event streams become more critical in low-light scenes or when severe motion blur degrades frame quality.

\section{Discussion} \label{sec:discussion}
Active event-based stereo vision offers a promising solution for high-speed depth perception, paving the way for next-generation depth cameras for agile robotics. Here, we will discuss the practicality and the limitations of our solution.

\textbf{\emph{Practicality}}. Although the proposed lightweight ActiveEventNet achieves high computational efficiency (e.g., 150 FPS on an NVIDIA 3090 GPU), its inference speed is limited by its synchronous frame-based processing paradigm, thus failing to match the microsecond-level temporal resolution of event cameras. Designing an asynchronous event-based paradigm capable of achieving ultra-low latency inference on resource-constrained mobile platforms remains an open challenge. In addition, the current stereo camera system serves as a simplified prototype primarily for data acquisition. There is significant potential for miniaturization, similar to the RealSense D400 series, enabling more compact and practical deployment. Although our prototype employs the DAVIS346 to provide both events and frames, the proposed system is inherently scalable to higher resolution event cameras in future implementations. We believe our solution provides a novel perspective for advancing 3D perception in agile robotic platforms.

\textbf{\emph{Limitations}}. All depth data from the RealSense camera used as ground truth are collected only under normal-speed motion. This design choice follows a common practice in existing event-based depth estimation datasets, as neither commodity RGB-D sensors nor LiDARs can reliably provide accurate depth labels in high-speed motion scenarios. In practice, this work evaluates depth sensing in real high-speed scenes only through qualitative visual assessment, as no quantitative ground truth labels are available for these scenarios. Additionally, the raw RealSense depth labels can be noisy and exhibit occlusions. Future iterations could benefit from data preprocessing strategies (e.g., temporal filtering) to generate higher-quality depth labels. In short, acquiring high-quality ground truth for depth estimation in high-speed scenes remains an open challenge.

\section{Conclusion} \label{sec:conclusion}
This paper presents a novel active event-based stereo vision paradigm, which encompasses a stereo camera prototype, new datasets, and lightweight matching models for low-latency depth estimation. To our knowledge, this work represents an early exploration into integrating binocular event cameras with an infrared 2D pattern projector for ultrafast dense depth sensing. To this end, we first establish a real-world dataset using our active event-based stereo camera prototype, along with a synthetic dataset. Then, we propose ActiveEventNet+, a lightweight yet effective event-based stereo matching neural network designed to generate high-quality dense disparity maps from stereo event streams with low latency. Our solution demonstrates superior depth sensing performance compared to conventional frame-based stereo cameras in low-light, low-texture, and high-speed motion scenarios. We believe that our camera prototype can provide new insights into the development of next-generation neuromorphic stereo camera systems.

\ifCLASSOPTIONcompsoc
  \section*{Acknowledgments}
\else
  \section*{Acknowledgment}
\fi
This work was supported in part by the National Natural Science Foundation of China under Grant 62506190 and Grant 61827804, in part by the National Key R\&D Program of China 2018AAA0102801.


\ifCLASSOPTIONcaptionsoff
  \newpage
\fi


\bibliographystyle{IEEEtran}
\bibliography{IEEEabrv, tpami_references}


\begin{IEEEbiography}[{\includegraphics[width=1in,height=1.0in,clip,keepaspectratio]{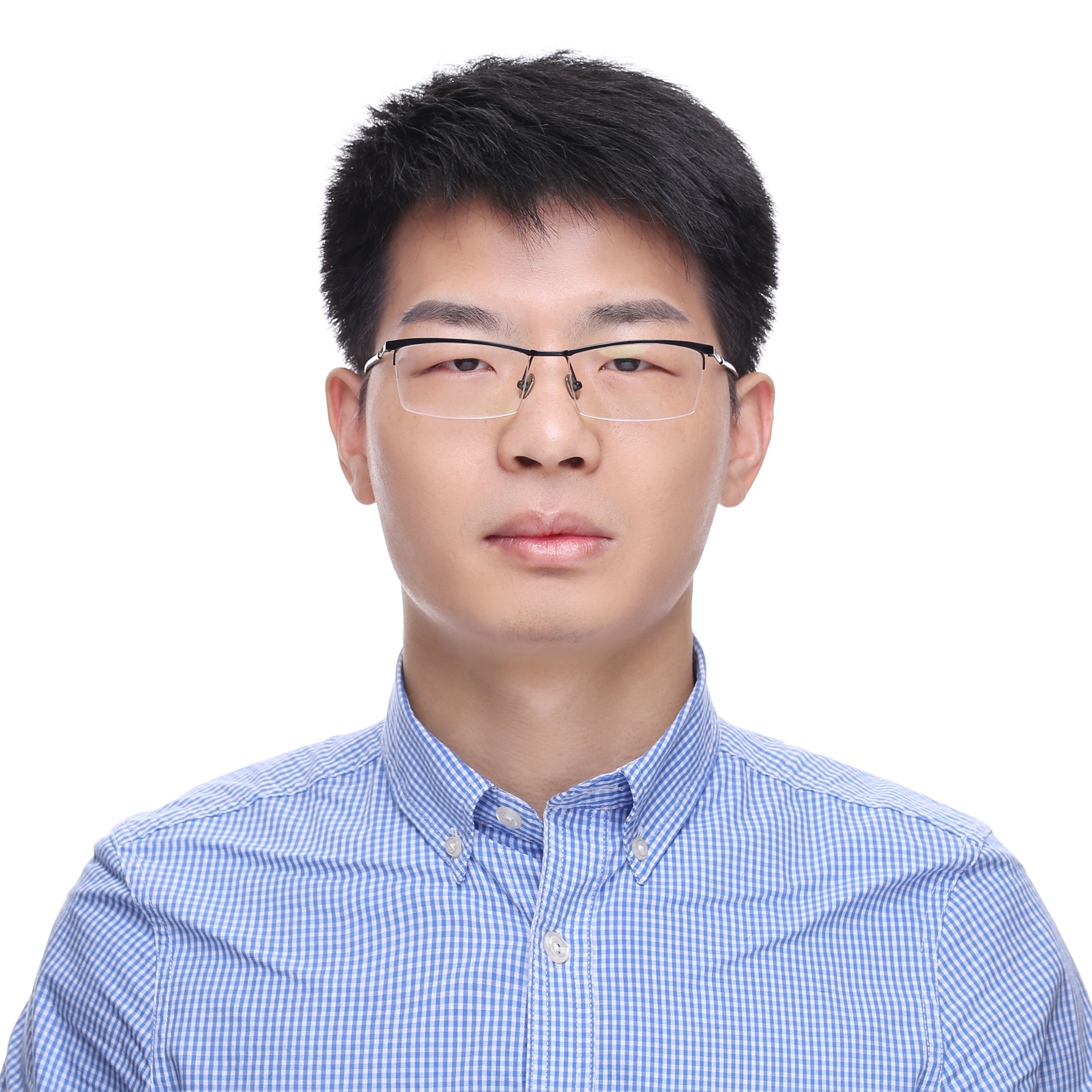}}]{Jianing Li} (Member,~IEEE) received the Ph.D. degree from the National Engineering Research Center for Visual Technology, School of Computer Science, Peking University, Beijing, China, in 2022. He is currently an associate researcher in the School of Computer Science at Peking University, Beijing, China. He is the author or coauthor of over 50 technical papers in refereed journals and conferences, such as TPAMI, TIP, CVRP, ICCV, NeurIPS, and ICML. He received the Outstanding Research Award from Peking University in 2020. He was honored with the Outstanding PhD Thesis Award from the CIE in 2024. His research interests include neuromorphic computing, event-based vision, neuromorphic engineering, and robot learning.
\end{IEEEbiography}

\begin{IEEEbiography}[{\includegraphics[width=1in,height=1.0in,clip,keepaspectratio]{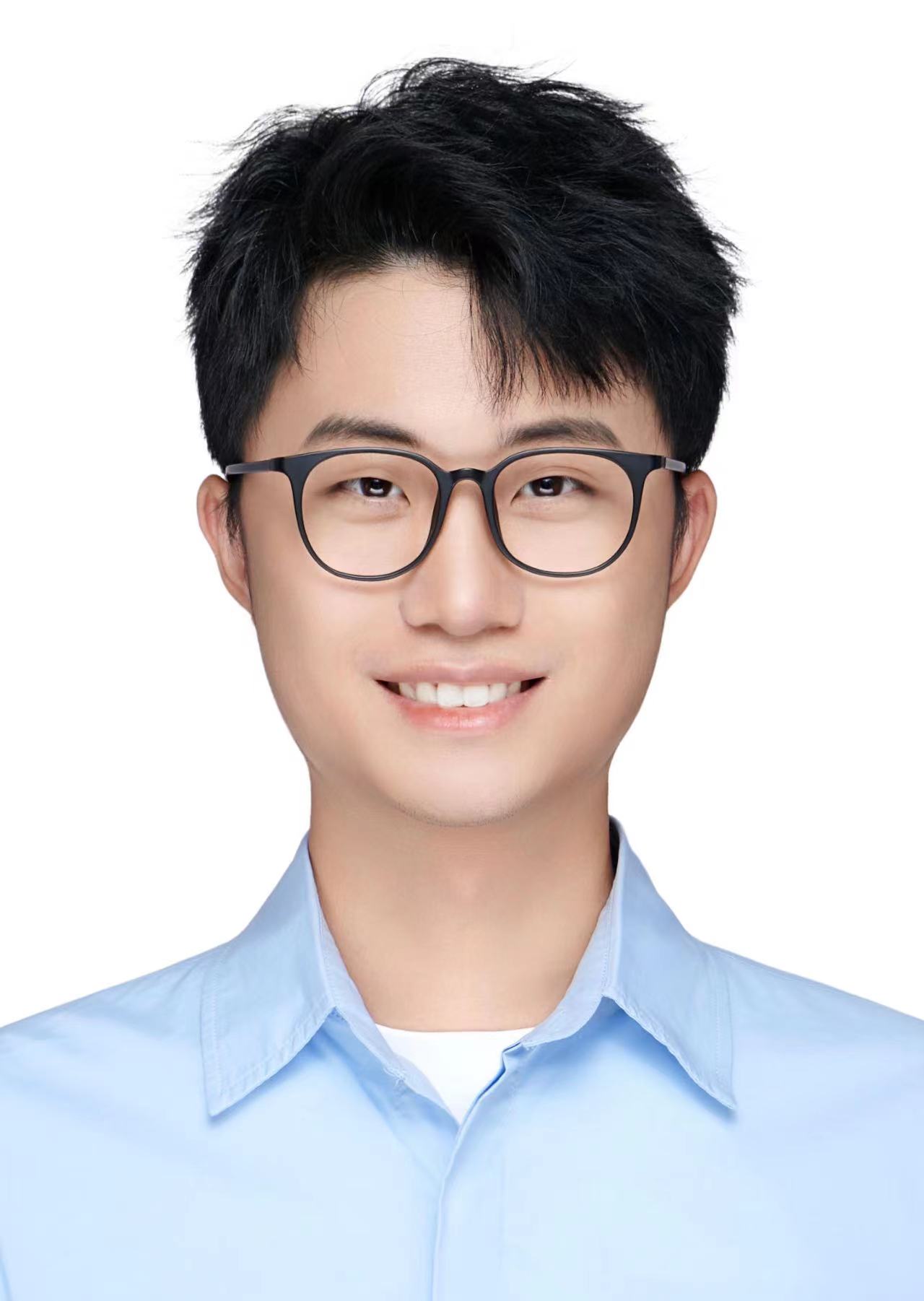}}]{Yunjian Zhang} received the B.S. degree from Northeastern University, China, in 2017, and the Ph.D. degree from the Institute of Information Engineering, Chinese Academy of Sciences, China, in 2023. His research interests include trustworthy AI and and 3D robotic vision.
\end{IEEEbiography}

\begin{IEEEbiography}[{\includegraphics[width=1in,height=1.0in,clip,keepaspectratio]{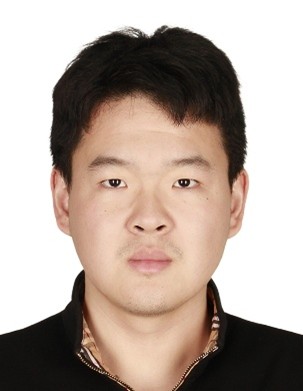}}]{Haiqian Han} is currently pursuing the the M.S. degree in the Department of Automation, Tsinghua University, Beijing, China, under the supervision of Prof. Xiangyang Ji. He received the B.E. degree in Robotics from Zhejiang University, Hangzhou, China, in 2023. His research interests include event-based vision, robotics, and computer vision.
\end{IEEEbiography}

\begin{IEEEbiography}[{\includegraphics[width=1.2in,height=1.2in,clip,keepaspectratio]{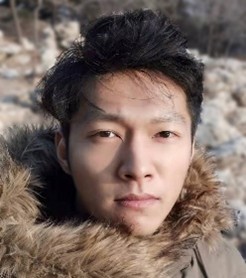}}]{Kangyao Huang} is currently pursuing the Ph.D degree in Computer Science at Tsinghua University, Beijing, China. He received the B.Eng. in Aerospace from Northwestern Polytechnical University, Xi’an, China, in 2016, and M.Res. in Control \& Systems Engineering from the University of Sheffield, Sheffield, U.K., in 2020. He has several years working experience in industry, and provided applied research in cooperation with partners in information, aerospace and manufacturing sectors. His research interests include robot learning and 3D vision.
\end{IEEEbiography}

\begin{IEEEbiography}[{\includegraphics[width=1in,height=1.0in,clip,keepaspectratio]{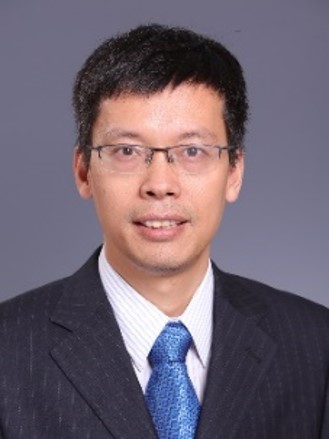}}]{Xiangyang Ji} (Member,~IEEE) received the BE degree in materials science and the MS degree in computer science from the Harbin Institute of Technology, Harbin, China, in 1999 and 2001, respectively, and the PhD degree in computer science from the Institute of Computing Technology, Chinese Academy of Sciences, Beijing, China. He joined Tsinghua University, Beijing, in 2008, where he is currently a professor with the Department of Automation, School of Information Science and Technology. He has authored more than 200 refereed conference and journal papers. His current research interests include signal processing, computer vision, and computational photography.
\end{IEEEbiography}

\end{document}